\documentclass[11pt]{article}

\usepackage{color}
\usepackage{graphicx}
\usepackage{algorithm}
\usepackage{algorithmic}
\usepackage{indentfirst}
\usepackage{amsfonts}
\usepackage{amsthm}
\usepackage{amsmath}
\usepackage{url}

\newcommand{\argmax}{\mathop{\rm arg~max}\limits}
\makeatletter
\newcommand{\figcaption}[1]{\def\@captype{figure}\caption{#1}}
\newcommand{\tblcaption}[1]{\def\@captype{table}\caption{#1}}
\def\Hline{
\noalign{\ifnum0=`}\fi\hrule \@height 2pt \futurelet
\reserved@a\@xhline}
\makeatother

\title{{\Large Understanding Community Structure in Layered Neural Networks}}

\newtheorem{theo}{Theorem}[section]

\begin{document}

\title{Understanding Community Structure \\in Layered Neural Networks}
\author{{\normalsize Chihiro Watanabe}\thanks{\textit{Email address:} watanabe.chihiro@lab.ntt.co.jp}{\normalsize ,\ \ Kaoru Hiramatsu,\ \ Kunio Kashino}\\
{\small\itshape NTT Communication Science Laboratories,}\\{\small\itshape 3-1, Morinosato Wakamiya, Atsugi-shi, Kanagawa Pref. 243-0198 Japan}}
\date{}

\maketitle

\begin{abstract}
A layered neural network is now one of the most common choices for the prediction or recognition of high-dimensional practical data sets, where the relationship 
between input and output data is complex and cannot be represented well by simple conventional models. 
Its effectiveness is shown in various tasks, such as image recognition and natural language processing, however, the lack of interpretability of the trained 
result by a layered neural network has limited its application area. 

In our previous studies, we proposed methods for extracting a simplified global structure of a trained layered neural network by applying a network 
analysis method and by classifying the units into communities according to their connection patterns with adjacent layers. 
These methods provided us with knowledge about the strength of the relationship between communities from the existence of bundled connections, which are 
determined by threshold processing of the connection ratio between pairs of communities. 

However, it has been difficult to understand the role of each community in detail or quantitatively by observing the resulting modular structure with these previous methods. 
We could only know to which sets of the input and output dimensions each community was mainly connected, by tracing the bundled connections from the community to 
the input and output layers. 
Another problem is that the finally obtained modular structure is changed greatly depending on the setting of the threshold hyperparameter used for determining bundled connections, 
leading to a different result to the discussion about the role of each community. 

In this paper, we propose a new method for interpreting quantitatively the role of each community in inference, which is extracted by our previous methods, 
by defining the effect of each input dimension on a community, and the effect of a community on each output dimension. 
We show experimentally that our proposed method can reveal the role of each part of a layered neural network by applying the neural networks to three types of 
data sets, extracting communities from the trained network, and applying the proposed method to the community structure. \vspace{0.5em}
\end{abstract}


\section{Introduction}

A layered neural network is now one of the most common machine learning models, and it has successfully improved the predictive performance for practical 
data sets in high-dimensional space, such as data sets of images (\cite{Krizhevsky2012,Tompson2014}), speech (\cite{Hinton2012,Sainath2013}) and natural language 
(\cite{Collobert2011,Sutskever2014}). 
It is a kind of stochastic model that is represented by hierarchically structured networks consisting of a large number of nonlinear parameters, which enable them to express 
the complex relationships between input and output values hidden in practical data sets. 
One major problem when using LNNs is that its prediction mechanism is black boxed, and it is difficult for human beings to understand how a trained LNN derives 
the map from input to output values. 
This issue of the interpretability of machine learning methods has begun to be viewed as a problem in recent years. 
For example, in the application areas of automatic driving and medical treatment, we cannot introduce a method if the reason for its predicition result cannot be explained. 
As examples of interpretable models, we can use linear models and decision trees instead of layered neural networks, however, with these simpler models it is difficult to 
capture the complex input-output relationships in practical data sets as well as layered neural networks. 

Therefore, to understand or obtain knowledge from a layered neural network, research on various approaches has been proposed in recent years. 
\begin{itemize}
\item \textbf{Approach A: Analysis of trained layered neural networks}
  \begin{itemize}
  \item Feature selection and approximation: This approach filters the important features that are mainly used for inference and then approximates the function 
  of the layered neural network at data points by using simple models such as a linear model or a decesion tree (\cite{Ribeiro2016,Lundberg2017,Nagamine2017}). 
  This approach attempts to interpret the function of a whole layered neural network for a given data set. 
  \item Analysis of unit outputs and their mutual relationships: This approach derives unit output information from a trained layered neural network 
  (\cite{Luo2016,Zahavy2016,Raghu2017}), such as the similarity of outputs for a given data set and the effect of each unit on the neural network output. 
  \item Analysis of the influence on neural network inference by data: This approach detects which part of a data sample or which features affect the prediction 
  result provided by a layered neural network (\cite{Adler2016,Balu2017,Koh2017,Kulynych2017,Zhang2017}). 
  \end{itemize}
\item \textbf{Approach B: Training interpretable layered neural networks}
  \begin{itemize}
  \item This approach is used to devise the training methods so that the trained layered neural network is represented by an interpretable function 
  (\cite{Foerster2017,Gonzalez2017}). For instance, training a network has been proposed as a way of performing an affine transformation or one that can be represented 
  in a rule-based manner. 
  \end{itemize}
\item \textbf{Other approaches}
  \begin{itemize}
  \item An approach designed to train another NN to provide an explanation for a prediction result provided by a layered neural network (\cite{Barratt2017}). 
  \item An approach designed to define the interpretability of each layer of a convolutional neural network by how the roles of the units are differentiated (\cite{Bau2017}). 
  More specifically, in an image segmentation task, it enumerates the number of image concepts that maximize the unit output in a layer. 
  \end{itemize}
\end{itemize}
The methods described in the above related studies have enabled us to acquire knowledge about the mechanism of a whole layered neural network or that of each unit in a layered 
neural network, however, they cannot capture the global network structure of a trained layered neural network. 

Therefore, we proposed the first method for extracting a global layered neural network structure by applying network analysis to a trained layered 
neural network (\cite{Watanabe2017d,Watanabe2017b,Watanabe2018,Watanabe2017a}). 
By using these methods, we can decompose units in a layered neural network into groups or \textit{communities}, where all units have similar connection patterns with adjacent layers. 
The extracted community structure of a layered neural network provides information about the roles of each part of a layered neural network. 

Some problems remain with our previous methods: A trained layered neural network structure is simplified by summarizing multiple connections between communities into a single 
\textit{bundled connection}, based on threshold processing of the connection ratio between communities. 
The resulting structure changes greatly depending on the hyperparameter setting of the threshold, and this leads to a different discussion result as regards 
the layered neural network mechanism. 
The optimal setting of this hyperparameter is unknown, and there is no method for interpreting the role of each extracted community quantitatively. 
In these studies, the role of each community is inferred by tracing the bundled connections to determine which input and output dimensions mainly connect with the 
community. However, with this method, we could not obtain detailed information about how much a community is affected by each input dimension or how much 
contribution a community makes to each output dimension. 

To solve the above problem, in this paper we propose a new method for interpreting the roles of each community in a quantitative way in terms of the 
strength of the relationship with each input and output dimension. Our new method analytically calculates the following values to quantify the role of each community: 
(a) the degree to which the change in each input dimension affects the outputs of units in the community, and 
(b) the degree to which the change in outputs in the community affects the outputs in the output layer. 

In section \ref{sec:method}, we describe in detail a method for analyzing the role of each community. The communities are extracted by applying network analysis to a 
trained layered neural network. Then, in section \ref{sec:experiments}, we perform experiments using both synthetic and practical data sets and show that the proposed method 
can successfully quantify the role of each part of a layered neural network. We discuss the experimental results in section \ref{sec:discussions}. 
Finally, we conclude this paper in section \ref{sec:conclusion}. 


\section{Detecting and Understanding Community Structure in Layered Neural Networks}
\label{sec:method}

In this section, we describe a way to train a layered neural network, extract the community structure from the trained layered neural network, and 
analyze the predictive role of each community. 

\subsection{Training Layered Neural Network Based on Error Back Propagation}

We use the same method as that used in our previous work (\cite{Watanabe2018}) for training layered neural networks. 
Let $(x, y)$, $x\in \mathbb{R}^M$, $y\in \mathbb{R}^N$ be a data set of input data $x$ and output data $y$. 
We assume a probability density function $q(x,y)$ on $\mathbb{R}^M\times \mathbb{R}^N$ for training and test data sets. 
We use a training data set $\{(X_n,Y_n)\}_{n=1}^{n_1}$ with sample size $n_1$ to train a layered neural network with a function $f(x,w)$ 
from $x\in \mathbb{R}^M,\ w\in \mathbb{R}^L$ to $\mathbb{R}^N$ that predicts output data $y$ from input data $x$ and a parameter $w$. 
A parameter $w=\{\omega^d_{ij}, \theta^d_i\}$ of a layered neural network with $D$ layers consists of the connection weight $\omega^d_{ij}$ 
between the $i$-th unit in the depth $d$ layer and the $j$-th unit in the depth $d+1$ layer, and the bias $\theta^d_i$ of the $i$-th unit in the depth $d$ layer. 
Here, the depth $1$ and $D$ layers correspond to the input and output layers, respectively. 
The function $f_j(x,w)$ of layered neural network for the $j$-th unit in the output layer is represented as follows. 
\begin{eqnarray*}
  f_j(x,w) = \sigma(\sum_i \omega^{D-1}_{ij} o^{D-1}_i+\theta^{D-1}_j),
\end{eqnarray*}
where
\begin{eqnarray*}
  o^{D-1}_j = \sigma(\sum_i \omega^{D-2}_{ij} o^{D-2}_i+\theta^{D-2}_j),\ \ \ \ \ \ \cdots, \ \ \ \ \ \ o^2_j = \sigma(\sum_i \omega^1_{ij} x_i +\theta^1_j),
\end{eqnarray*}
and
\begin{eqnarray*}
  \sigma (x)=\frac{1}{1+\exp (-x)}.
\end{eqnarray*}

The training error $E(w)$ and the generalization error $G(w)$, respectively, are given by
\begin{eqnarray*}
  E(w) &=& \frac{1}{n_1} \sum_{n=1}^{n_1} \|Y_n-f(X_n,w)\|^2,\\
  G(w) &=& \int \|y-f(x,w)\|^2 q(x,y)dxdy,
\end{eqnarray*}
where $\|\cdot\|$ is the Euclidean norm of $\mathbb{R}^N$. Our goal is to minimize the generalization error, however, it cannot be calculated by a dataset with 
a finite sample size. Therefore, we approximate the generalization error by 
\begin{eqnarray*}
  G(w)\approx \frac{1}{m_1} \sum_{m=1}^{m_1} \|{Y_m}'-f({X_m}',w)\|^2,
\end{eqnarray*}
where $\{({X_m}', {Y_m}')\}_{m=1}^{m_1}$ is a test data set that is independent of the training data set. 

Layered neural networks usually have many parameters, leading to overfitting to a given training data set. 
To avoid overfitting, we adopt the LASSO method (\cite{Ishikawa1990,Tibshirani1996}), where the objective function to be minimized is given by the following $H(w)$. 
\begin{eqnarray*}
  H(w) &=& \frac{n_1}{2}\ E(w)+\lambda \sum_{d,i,j} |\omega^d_{ij}|,
\end{eqnarray*}
where $\lambda$ is a hyperparameter. 
With this method, we can obtain sparse connections, by deleting connection weights with small absolute values. 
The value of this objective function $H(w)$ is minimized with the stochastic steepest descent method, 
\begin{eqnarray}
  \Delta w = -\eta \nabla H_n (w)= -\eta \ \Bigl(\frac{1}{2}\nabla \{\|Y_n-f(X_n,w)\|^2\}+\lambda \ \mathrm{sgn}(w)\Bigr),
  \label{eq:dw}
\end{eqnarray}
where $H_n (w)$ is the training error computed solely from the randomly chosen $n$-th sample $(X_n, Y_n)$, and $\eta$ for training time $t$ is defined such that 
$\eta(t)\propto 1/t$. 
In this paper, we used the definition $\eta =0.7\times a_1 n_1/(a_1 n_1 +5t)$, where $a_1$ is the mean iteration number of layered neural network training per dataset. 
To numerically calculate Equation (\ref{eq:dw}), we used the following algorithm called error back propagation (\cite{Werbos1974,Rumelhart1986}). 
For the $D$-th layer, 
\begin{eqnarray*}
  \delta^{D}_j &=& (o^{D}_j-y_j)\ (o^{D}_j\ (1-o^{D}_j) +\epsilon_1),\\
  \Delta \omega^{D-1}_{ij} &=& -\eta (\delta^{D}_j o^{D-1}_i+\lambda \ \mathrm{sgn}(\omega^{D-1}_{ij})),\\
  \Delta \theta^D_j &=& -\eta \delta^{D}_j.
\end{eqnarray*}
For $d=D-1,\ D-2,\cdots, 2$,
\begin{eqnarray*}
  \delta^d_j &=& \sum_{k=1}^{l_{d+1}} \delta^{d+1}_k \omega^d_{jk}\ (o^d_j\ (1-o^d_j) +\epsilon_1),\\
  \Delta \omega^{d-1}_{ij} &=& -\eta (\delta^d_j o^{d-1}_i+\lambda \ \mathrm{sgn}(\omega^{d-1}_{ij})),\\
  \Delta \theta^d_j &=& -\eta \delta^d_j. 
\end{eqnarray*}
Here, $\epsilon_1$ is a hyperparameter for the convergence of a layered neural network. 
By iteratively calculating the above equations with randomly chosen training data samples $(X_n, Y_n)$, we can train the parameter of a layered neural network. 


\subsection{Community Detection from Trained Layered Neural Networks}

To improve the interpretability of trained layered neural networks, we have proposed decomposing their units into communities according to the network 
connection patterns. 
In our previous work (\cite{Watanabe2018, Watanabe2017a, Watanabe2017b, Watanabe2017d}), we used community detection models that cannot be applied when 
there are any units with no connections from or to adjacent layers. 
To extend the model for application to such cases, in this paper, we propose a probabilistic model for community detection that extends the APBEMA model 
(\cite{Wang2008}) to layered neural networks. 
The units in each layer of a neural network have connections only with the units in the two adjacent layers. 
Therefore, community detection from a layered neural network trained with a data set is accomplished layer-wise by observing the connection patterns with the 
adjacent layers. 

Here, we focus on an arbitrary layer in a trained neural network and explain the method for detecting communities in the layer. 
Let $u=\{u_k\}$, $t=\{t_i\}$, and $v=\{v_j\}$ be the units in the focused layer, namely those in the input-side and output-side adjacent layers, respectively. 
The connection patterns between input-side and output-side adjacent layers can be represented by adjacency matrices $A^+,\ A^-,\ B^+,\ B^-$: 
We use the definition $A^+_{i,k}=1$ if the connection weight is the threshold value $\xi$ or larger between the unit $u_k$ and the unit $t_i$ in the input-side layer, and 
$A^+_{i,k}=0$ otherwise. Similarly, we define $A^-_{i,k}=1$ if the connection weight between those units is the threshold value $-\xi$ or smaller, and 
$A^-_{i,k}=0$ otherwise. 
In the same way, we define the values of $B^+_{k,j}$ and $B^-_{k,j}$, by the connection weight between the unit $u_k$ and the unit $v_j$ in the output-side layer. 
We use the adjacency matrices $A^+,\ A^-,\ B^+,\ B^-$ as the observed data for classifying the unit set $u$. 

Let $\pi_c$ be the probability that a unit belongs to the community $c$. The parameter $\pi_c$ satisfies the following condition: 
\begin{eqnarray*}
  \sum_c \pi_c=1.
  \label{eq:normalization}
\end{eqnarray*}
Let $\tau^+_{i,k}$ and $\tau^-_{i,k}$, respectively, be the probabilities that a unit in the community $c$ has a positive and a negative connection weight with 
the unit $t_i$ in the input-side layer. 
In the same way, we define the parameters $\tau'^+_{k,j}$ and $\tau'^-_{k,j}$, respectively, as the probabilities that a unit in the community $c$ has a 
positive and a negative connection weight with the unit $v_j$ in the output-side layer. 
By introducing a hidden variable $g_k$, which represents the community of unit $u_k$, we assume that the probability of the adjacency matrices 
$A^+,\ A^-,\ B^+,\ B^-$ and hidden variable set $g=\{g_k\}$ is given by 
\begin{eqnarray*}
  \lefteqn{ \mathrm{Pr}(A^+,A^-,B^+,B^-,g|\pi,\tau^+,\tau^-,\tau'^+,\tau'^-)}\\
  &&=\mathrm{Pr}(A^+,A^-,B^+,B^-|g,\pi,\tau^+,\tau^-,\tau'^+,\tau'^-)\ \mathrm{Pr}(g|\pi,\tau^+,\tau^-,\tau'^+,\tau'^-),
\end{eqnarray*}
where 
\begin{eqnarray*}
  \lefteqn{ \mathrm{Pr}(A^+,A^-,B^+,B^-|g,\pi,\tau^+,\tau^-,\tau'^+,\tau'^-) }\\
  &&=\prod_k \Bigl\{ \prod_i {\Bigl(\tau^+_{g_k,i}\Bigr)}^{A^+_{i,k}} {\Bigl(1-\tau^+_{g_k,i}\Bigr)}^{1-A^+_{i,k}} 
  {\Bigl(\tau^-_{g_k,i}\Bigr)}^{A^-_{i,k}} {\Bigl(1-\tau^-_{g_k,i}\Bigr)}^{1-A^-_{i,k}} \Bigr\} \\
  &&\Bigl\{ \prod_j {\Bigl(\tau'^+_{g_k,j}\Bigr)}^{B^+_{k,j}} {\Bigl(1-\tau'^+_{g_k,j}\Bigr)}^{1-B^+_{k,j}} 
  {\Bigl(\tau'^-_{g_k,j}\Bigr)}^{B^-_{k,j}} {\Bigl(1-\tau'^-_{g_k,j}\Bigr)}^{1-B^-_{k,j}} 
  \Bigr\},
  \label{eq:gAB}
\end{eqnarray*}
and
\begin{eqnarray}
  \mathrm{Pr}(g|\pi,\tau^+,\tau^-,\tau'^+,\tau'^-) = \prod_k \pi_{g_k}.
  \label{eq:gg}
\end{eqnarray}

Therefore, the log likelihood $\mathcal{L}$ of $A^+,A^-,B^+,B^-$ and $g$ is given by
\begin{eqnarray*}
  \mathcal{L} &\equiv& \ln \mathrm{Pr}(A^+,A^-,B^+,B^-,g|\pi,\tau^+,\tau^-,\tau'^+,\tau'^-)\\
  &=& \sum_k \left\{ \ln \pi_{g_k} +\sum_i (A^+_{i,k} \ln \tau^+_{g_k,i} +(1-A^+_{i,k}) \ln (1-\tau^+_{g_k,i})\right. \\
  &&+A^-_{i,k} \ln \tau^-_{g_k,i} +(1-A^-_{i,k}) \ln (1-\tau^-_{g_k,i}))\\
  &&+ \sum_j (B^+_{k,j} \ln \tau'^+_{g_k,j} +(1-B^+_{k,j}) \ln (1-\tau'^+_{g_k,j})\\
  &&\left. +B^-_{k,j} \ln \tau'^-_{g_k,j} +(1-B^-_{k,j}) \ln (1-\tau'^-_{g_k,j})) \right\}.
\end{eqnarray*}

The expected value $\bar{\mathcal{L}}$ of the above log likelihood over the hidden variable set $g$ is given by
\begin{eqnarray*}
  \bar{\mathcal{L}} = \sum_g \mathrm{Pr} (g|A^+,A^-,B^+,B^-,\pi,\tau^+,\tau^-,\tau'^+,\tau'^-) \mathcal{L} \\
\end{eqnarray*}
\begin{eqnarray*}
  &=& \sum_{k,c} q_{k,c} \left\{ \ln \pi_c +\sum_i (A^+_{i,k} \ln \tau^+_{c,i} +(1-A^+_{i,k}) \ln (1-\tau^+_{c,i})\right. \\
  &&+A^-_{i,k} \ln \tau^-_{c,i} +(1-A^-_{i,k}) \ln (1-\tau^-_{c,i}))\\
  &&+ \sum_j (B^+_{k,j} \ln \tau'^+_{c,j} +(1-B^+_{k,j}) \ln (1-\tau'^+_{c,j})\\
  &&\left. +B^-_{k,j} \ln \tau'^-_{c,j} +(1-B^-_{k,j}) \ln (1-\tau'^-_{c,j})) \right\},
  \label{eq:exploglh}
\end{eqnarray*}
Here, the variable $q_{k,c}$ represents the probability that the unit $u_k$ belongs to the community $c$, and it is defined by the following equation: 
\begin{eqnarray}
  \lefteqn{ q_{k,c}\equiv \mathrm{Pr}(g_k=c|A^+,A^-,B^+,B^-,\pi,\tau^+,\tau^-,\tau'^+,\tau'^-) }\nonumber \\
  &&= \frac{\mathrm{Pr}(A^+,A^-,B^+,B^-,g_k=c|\pi,\tau^+,\tau^-,\tau'^+,\tau'^-)}{\mathrm{Pr}(A^+,A^-,B^+,B^-|\pi,\tau^+,\tau^-,\tau'^+,\tau'^-)}.
  \label{eq:qdefine}
\end{eqnarray}

The optimal parameters $\pi,\tau^+,\tau^-,\tau'^+,\tau'^-$ and the variable $q_{k,c}$ for maximizing $\bar{\mathcal{L}}$ of Equation (\ref{eq:exploglh}) are 
iteratively estimated with the EM algorithm, which is represented by the following Equations (\ref{eq:q}) and (\ref{eq:pitau}). 

\begin{theo}
If $\{q_{k,c}\},\ \{\pi_c\},\ \{\tau^+_{c,i}\},\ \{\tau^-_{c,i}\},\ \{\tau'^+_{c,j}\},\ \{\tau'^-_{c,j}\}$ maximizes $\bar{\mathcal{L}}$, then they satisfy 
\begin{eqnarray}
  q_{k,c}=\frac{r_{k,c}}{\sum_s r_{k,s}},
  \label{eq:q}
\end{eqnarray}
and
\begin{eqnarray}
  \lefteqn{ \pi_c = \frac{\sum_k q_{k,c}}{k_0}, \ \ \ \tau^+_{c,i} = \frac{\sum_k A^+_{i,k} q_{k,c}}{\sum_k q_{k,c}}, \ \ \ 
  \tau^-_{c,i} = \frac{\sum_k A^-_{i,k} q_{k,c}}{\sum_k q_{k,c}}, }\nonumber \\
  &&\tau'^+_{c,j} = \frac{\sum_k B^+_{k,j} q_{k,c}}{\sum_k q_{k,c}}, \ \ \ 
  \tau'^-_{c,j} = \frac{\sum_k B^-_{k,j} q_{k,c}}{\sum_k q_{k,c}},
  \label{eq:pitau}
\end{eqnarray}
where
\begin{eqnarray*}
  \lefteqn{ r_{k,c}\equiv \pi_c \left[ \prod_i \Bigl(\tau^+_{c,i}\Bigr)^{A^+_{i,k}} \Bigl(1-\tau^+_{c,i}\Bigr)^{1-A^+_{i,k}} 
  \Bigl(\tau^-_{c,i}\Bigr)^{A^-_{i,k}} \Bigl(1-\tau^-_{c,i}\Bigr)^{1-A^-_{i,k}} \right] }\\
  &&\left[ \prod_j \Bigl(\tau'^+_{c,j}\Bigr)^{B^+_{k,j}} \Bigl(1-\tau'^+_{c,j}\Bigr)^{1-B^+_{k,j}} 
  \Bigl(\tau'^-_{c,j}\Bigr)^{B^-_{k,j}} \Bigl(1-\tau'^-_{c,j}\Bigr)^{1-B^-_{k,j}} \right].
\end{eqnarray*}
\end{theo}
Here, variable $k_0$ represents the number of elements in unit set $u$.

We show the proof of the above theorem in Appendix. A. 
From the above theorem, the optimal parameters $\pi, \tau^+,\tau^-,\tau'^+,\tau'^-$ for the given probability of community assignment and 
the probability of community assignment $q$ for the optimized parameters are iteratively estimated based on Equations. (\ref{eq:q}) and (\ref{eq:pitau}). 
The community of the unit $u_k$ in the center layer is determined by $\argmax_c q_{k,c}$.


\subsection{Understanding Community Structure in Layered Neural Networks}

In the previous sections, we described how we obtained the community structure of a trained layered neural network, however, no method has been reported for understanding 
the role of each community in a quantitative way. 
To analyze what is done by each part of the layered neural network, we propose a new method that defines the role of community $c$ as a pair of feature vectors 
$v^{\mathrm{in}}_{c}=\{v^{\mathrm{in}}_{ic}\}_{i=1,\cdots,i_0}$ and $v^{\mathrm{out}}_{c}=\{v^{\mathrm{out}}_{cj}\}_{j=1,\cdots,j_0}$. 
Here, $i_0$ and $j_0$ represent the number of input and output dimensions, respectively. 

By our definition the element of feature vector $v^{\mathrm{in}}_{ic}$ represents the effect of the $i$-th input dimension on the community $c$. 
The value of this element is calculated as the magnitude of the output fluctuations of the units in the community $c$ when the layered neural network cannot 
use the information of the $i$-th input dimension. 
More specifically, it is computed as the square root errors of the outputs in the community $c$, when the value of the $i$-th input dimension is replaced 
with the mean value for the training data, whichever data are input. 
This definition is given by the following equations. 
\\
\textbf{Effect of $i$-th input dimension on community $c$}: 
For the depths $d=2,\cdots,D$, let $o^{(n)}_k$ be the output of the $k$-th unit for the $n$-th input data sample $X^{(n)}$, 
and let $z^{(n)}_k$ be the output of the $k$-th unit for an input data sample $X'^{(n)}$ that is generated based on the following definition: 
\begin{eqnarray*}
  X'^{(n)}_i \equiv \frac{1}{n_1} \sum_n X^{(n)}_i. \\
  \mathrm{For\ } l\neq i,\ X'^{(n)}_l \equiv X^{(n)}_l. 
\end{eqnarray*}
We define $v^{\mathrm{in}}_{ic}=\sqrt{\frac{1}{n_1} \sum_{k\in u(c)} \sum_n \Bigl( o^{(n)}_k -z^{(n)}_k \Bigr)^2}$, where $u(c)$ represents the set of all the units in the community $c$. 

Similarly, we define the element of feature vector $v^{\mathrm{out}}_{cj}$ as representing the effect of the community $c$ on the $j$-th output dimension. 
The value of this element is calculated as the magnitude of the output fluctuations of the $j$-th output dimension when the layered neural network cannot 
use the information of the outputs in the community $c$. 
This is computed as the square root error of the value of the $j$-th output dimension when the outputs in the community $c$ are replaced with the mean values 
for the training data, whichever data are input. 
This definition is given by the following equations. 
\\
\textbf{Effect of community $c$ on $j$-th output dimension}: 
For the depths $d=1,\cdots,D-1$, let $y^{(n)}_j$ be the output of the $j$-th unit in the output layer for the $n$-th input data sample $X^{(n)}$, 
and let $z^{(n)}_j$ be the output of the $j$-th unit in the output layer when changing the output values in the $d$-th layer, according to the following procedure: 
In the $d$-th layer, we change the output value of the $k$-th unit for the $n$-th input data sample from $o^{(n)}_k$ to $o'^{(n)}_k$. Here, $o'^{(n)}_k$ is given by 
\begin{eqnarray*}
  \mathrm{For\ } k\in u(c),\ o'^{(n)}_k \equiv \frac{1}{n_1} \sum_n o^{(n)}_k. \\
  \mathrm{For\ } k\not\in u(c),\ o'^{(n)}_k \equiv o^{(n)}_k. 
\end{eqnarray*}
We employ the definition $v^{\mathrm{out}}_{cj}=\sqrt{\frac{1}{n_1} \sum_n \Bigl( y^{(n)}_j -z^{(n)}_j \Bigr)^2}$. 

By using the above feature vectors $v^{\mathrm{in}}_{c}$ and $v^{\mathrm{out}}_{c}$, we can gain knowledge about the role of the community $c$ 
in terms of the degree to which it is affected by the value of each input dimension, and the degree to which it affects the value of each output dimension. 
In our previous studies (\cite{Watanabe2018, Watanabe2017a, Watanabe2017b, Watanabe2017d}), we proposed a further simplification of the community structure by 
defining \textit{bundled connections}, which summarize multiple connections between communities based on threshold processing. 
Compared with such previous methods, our new method has the merits that it can represent the role of each community quantitatively and in detail, 
and that the obtained results do not depend on the settings of any hyperparameters, such as the threshold hyperparameter used to define bundled 
connections in previous studies. 


\section{Experiments}
\label{sec:experiments}

Here, we show experimentally that our proposed method can explain the role of each community of a layered neural network by using both synthetic and practical data sets. 
We describe the detailed settings for the experiments in Appendix B. 

\subsection{Preliminary Experiment Using Synthetic Data Set}
\label{sec:exp1}

First, we applied our proposed method to a layered neural network that had been trained with a synthetic data set with a ground truth modular structure and confirmed whether or not 
the proposed method could properly show the relationship between each community and the input or output dimensions. 

We assume that the ground truth modular structure consists of three independent layered neural networks, all of which have the same architecture (number of units in a layer and 
number of hidden layers). In other words, there are three ground truth communities in each layer of the whole layered neural network. 
Each independent layered neural network is assumed to have $15$ units per layer and two hidden layers. 
The ground truth parameters $\hat{w}=\{\hat{\omega}^d_{ij}, \hat{\theta}^d_i\}$ of a layered neural network are generated by: 
\begin{eqnarray*}
  &&\hat{\omega}^d_{ij}\overset{\text\small\rm{i.i.d.}}{\sim}\mathcal{N}(0,1), \\
  &&\hat{\theta}^d_i\overset{\text\small\rm{i.i.d.}}{\sim}\mathcal{N}(0,0.5).
\end{eqnarray*}
Here, connection weights with absolute values of one or smaller were deleted. 
By using the above parameters, the training data set $\{(X_n,Y_n)\}$ was generated by: 
\begin{eqnarray*}
  &&X_n\overset{\text\small\rm{i.i.d.}}{\sim}\mathcal{N}(0,3), \\
  &&Y_n=f(X_n,\hat{w})+\epsilon_2,\ \ \ \ \ \ \epsilon_2 \overset{\text\small\rm{i.i.d.}}{\sim}\mathcal{N}(0,0.05).
\end{eqnarray*}

\begin{figure}[t]
  \hspace{-5mm}\includegraphics[width=140mm]{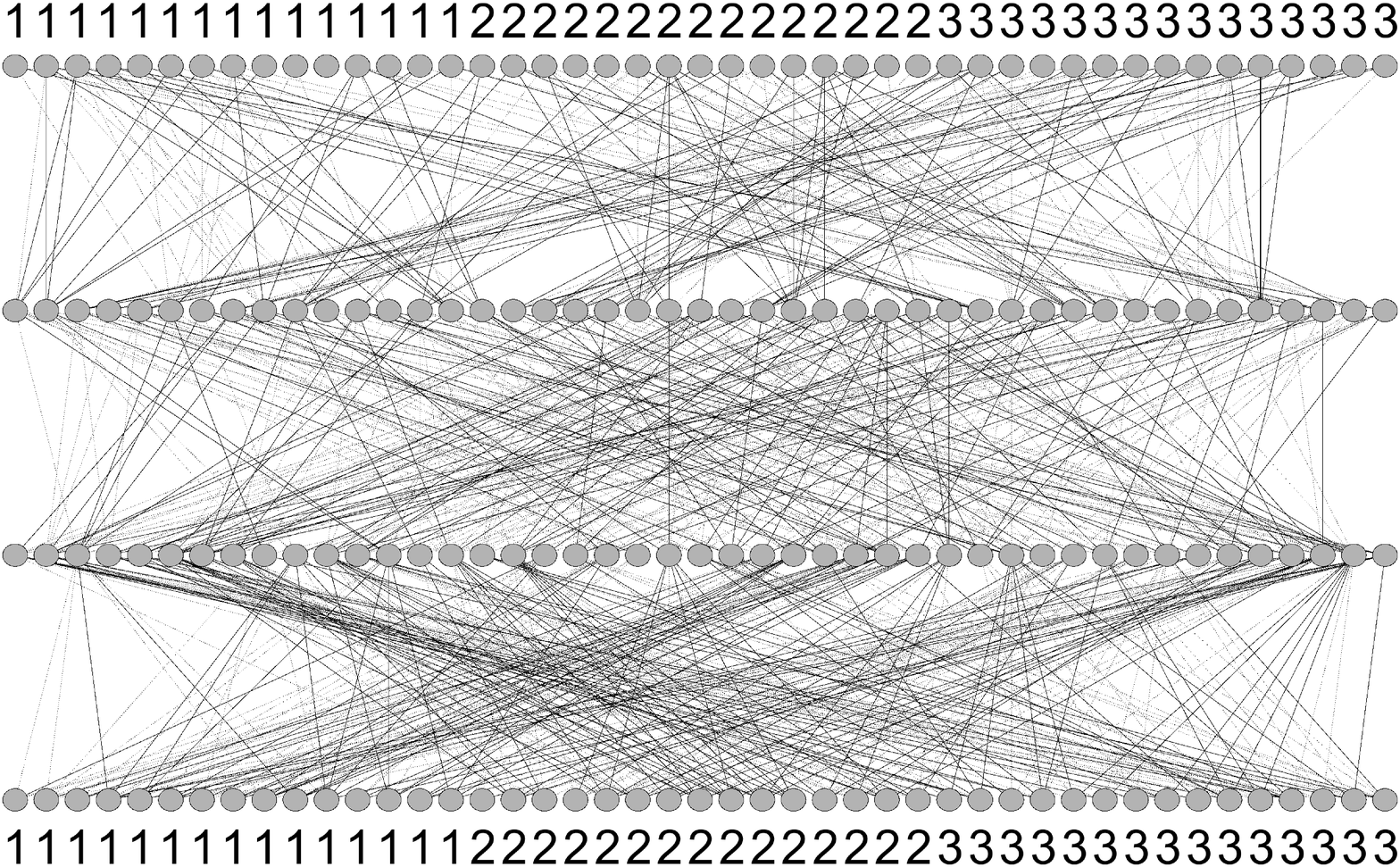}
  \caption{A layered neural network trained with a synthetic data set. 
  The top and bottom layers, respectively, correspond to the output and input layers. 
  Solid and dotted lines, respectively, represent the positive and negative connection weights. 
  The numbers $1$, $2$, and $3$ written next to the input and output units show the community indices in the ground truth structure. }\vspace{5mm}
  \label{fig:lnn_syn}
\end{figure}

\begin{figure}[t]
  \hspace{-5mm}\includegraphics[width=140mm]{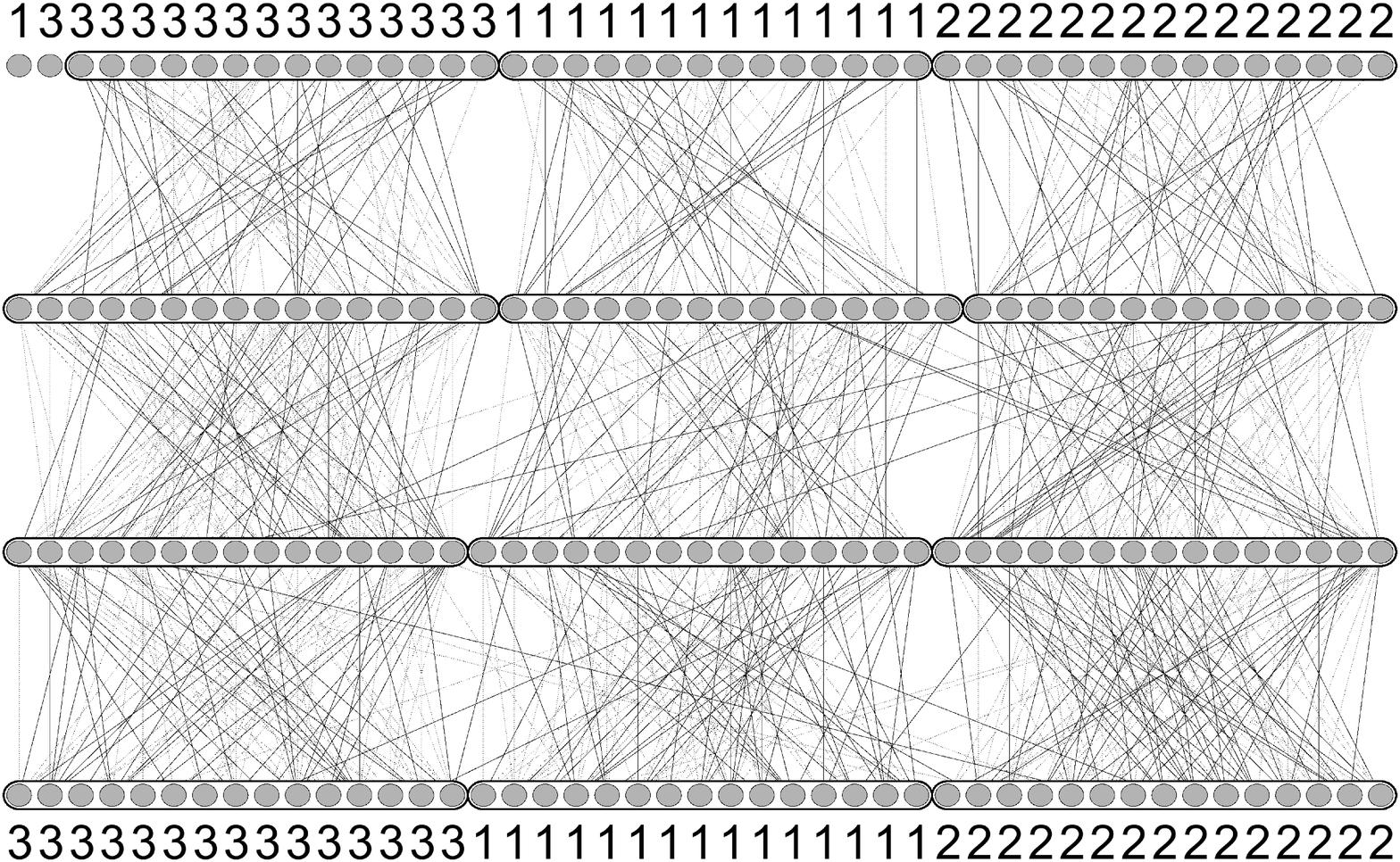}
  \caption{Community structure extracted from the trained layered neural network. 
  In each layer, the units were decomposed into three communities. }
  \label{fig:community_syn}
\end{figure}

\begin{figure}[t]
  \hspace{-15mm}\includegraphics[width=160mm]{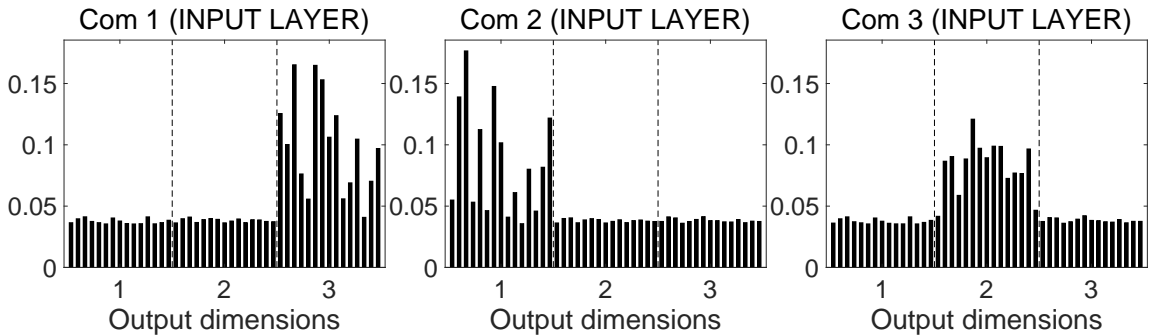}
  \caption{Effect of each community in the input layer on the output dimensions. 
  Com $1$, $2$, and $3$ represent the communities on the left, middle, and right in Figure \ref{fig:community_syn}. 
  The labels on the horizontal axis show the ground truth community indices of the input dimensions. 
  This figure shows that each community in the input layer corresponds to one of the three independent layered neural networks in the ground truth structure, 
  from the standpoint that it contributes mainly to the output dimensions of one community in the ground truth structure. }
  \label{fig:role_syn1}
\end{figure}

\begin{figure}[t]
  \hspace{-15mm}\includegraphics[width=160mm]{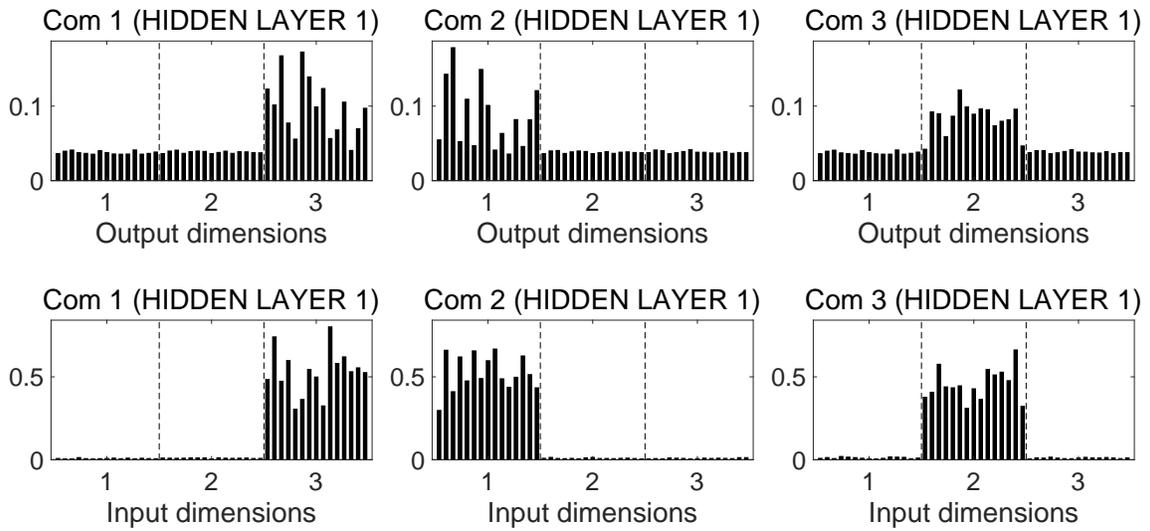}
  \caption{Top: Effect of each community in hidden layer $1$ (or the hidden layer adjacent to the input layer) on the output dimensions. 
  Bottom: Effect of the input dimensions on each community in hidden layer $1$. 
  As with Figure \ref{fig:role_syn1}, this figure shows that each community in hidden layer $1$ corresponds to one of the three independent neural networks. }
  \label{fig:role_syn2}
\end{figure}

\begin{figure}[t]
  \hspace{-15mm}\includegraphics[width=160mm]{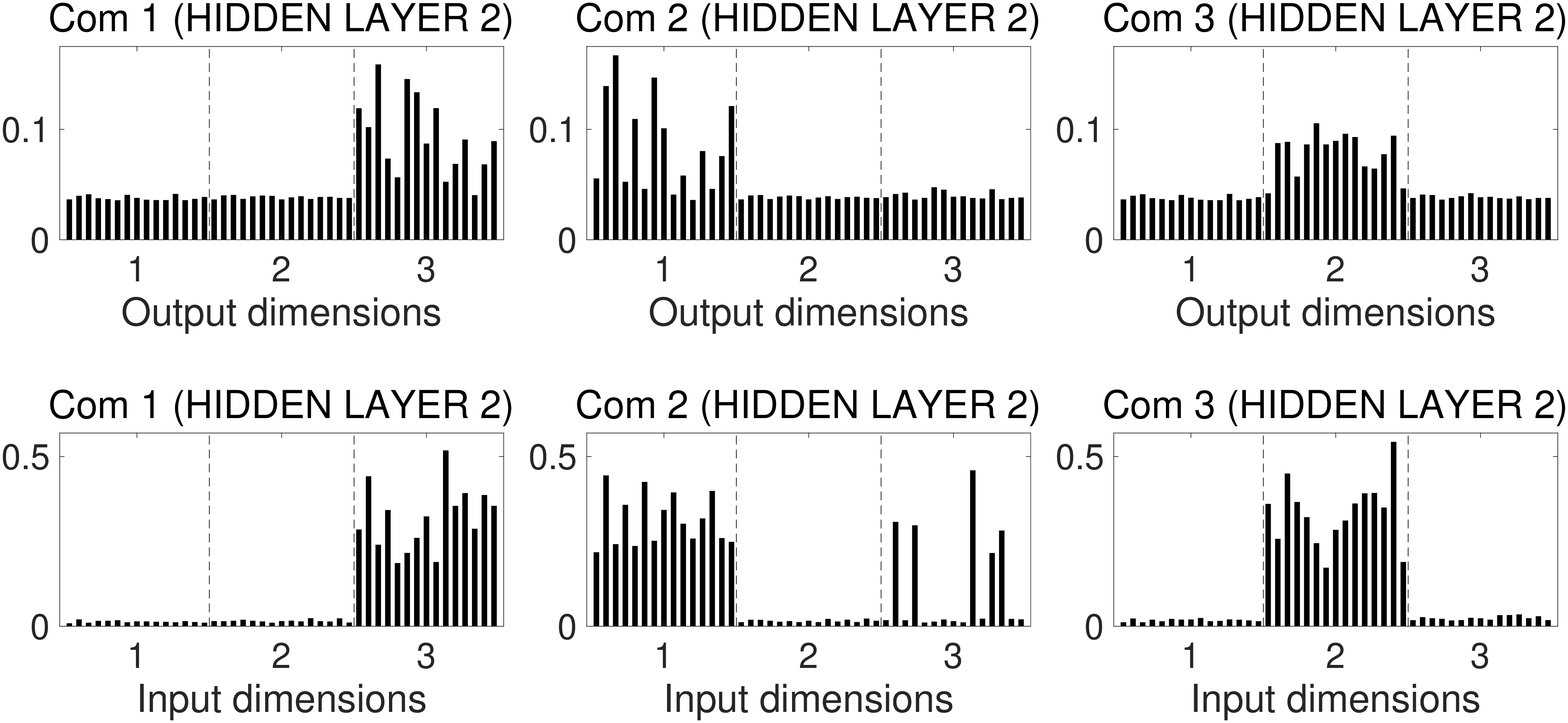}
  \caption{Top: Effect of each community in hidden layer $2$ (or the hidden layer adjacent to the output layer) on the output dimensions. 
  Bottom: Effect of the input dimensions on each community in hidden layer $2$. 
  in the ground truth structure. }
  \label{fig:role_syn3}
\end{figure}

\begin{figure}[t]
  \hspace{-15mm}\includegraphics[width=160mm]{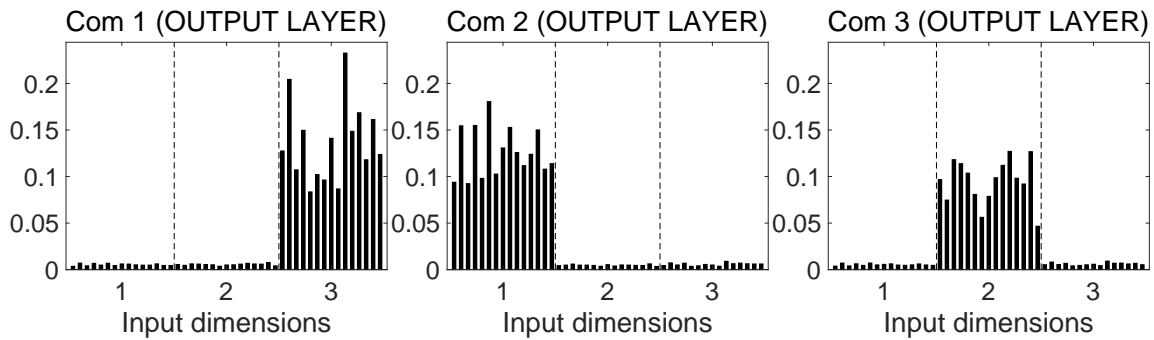}
  \caption{Effect of the input dimensions on each community in the output layer. }
  \label{fig:role_syn4}
\end{figure}

We train a layered neural network with the above data set $\{(X_n,Y_n)\}$, detected communities in the resulting trained network, and applied the proposed method. 
Figures \ref{fig:lnn_syn} and \ref{fig:community_syn} show the trained layered neural network, and the extracted community structure. 
Figures \ref{fig:role_syn1}, \ref{fig:role_syn2}, \ref{fig:role_syn3}, and \ref{fig:role_syn4} show the role of each community in the input, hidden and output 
layers, respectively. 

As shown in Figure \ref{fig:community_syn}, the units in each layer are decomposed into three communities, and there are more connections between the communities 
on the left, in the middle, and on the right side, than between other pairs of communities. This shows that the layered neural network was trained 
as a ground truth structure that consisted of three independent networks. 
Figures \ref{fig:role_syn1} to \ref{fig:role_syn4} provide more quantative information about the role of each community. It shows that each community corresponds 
to one of the three independent neural networks of the ground truth structure, in that it has relationships only with units in the same community 
in the input and output layer. These results show that our new proposed method can properly represent the role of each detected community when a 
ground truth modular structure is hidden in a data set. 


\subsection{Experiment Using Sequential Data Set of Consumer Price Index}
\label{sec:exp2}

\begin{figure}[t]
  \hspace{-15mm}\includegraphics[width=160mm]{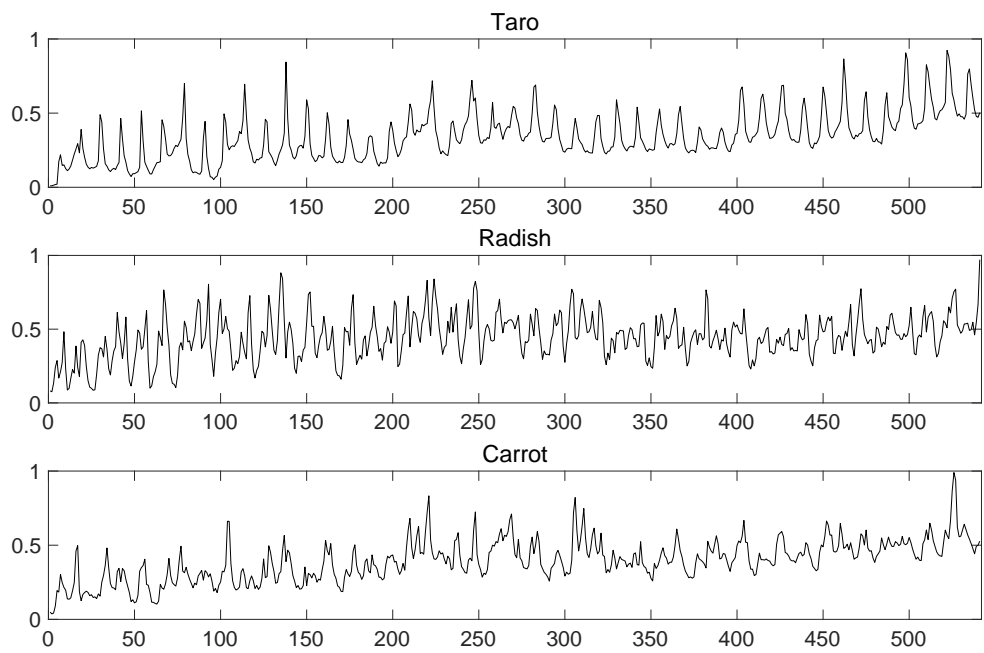}
  \caption{Output data of consumer price indices for taro, radish and carrot for $541$ months.}
  \label{fig:data_food}
\end{figure}

\begin{figure}[t]
  \hspace{-15mm}\includegraphics[width=160mm]{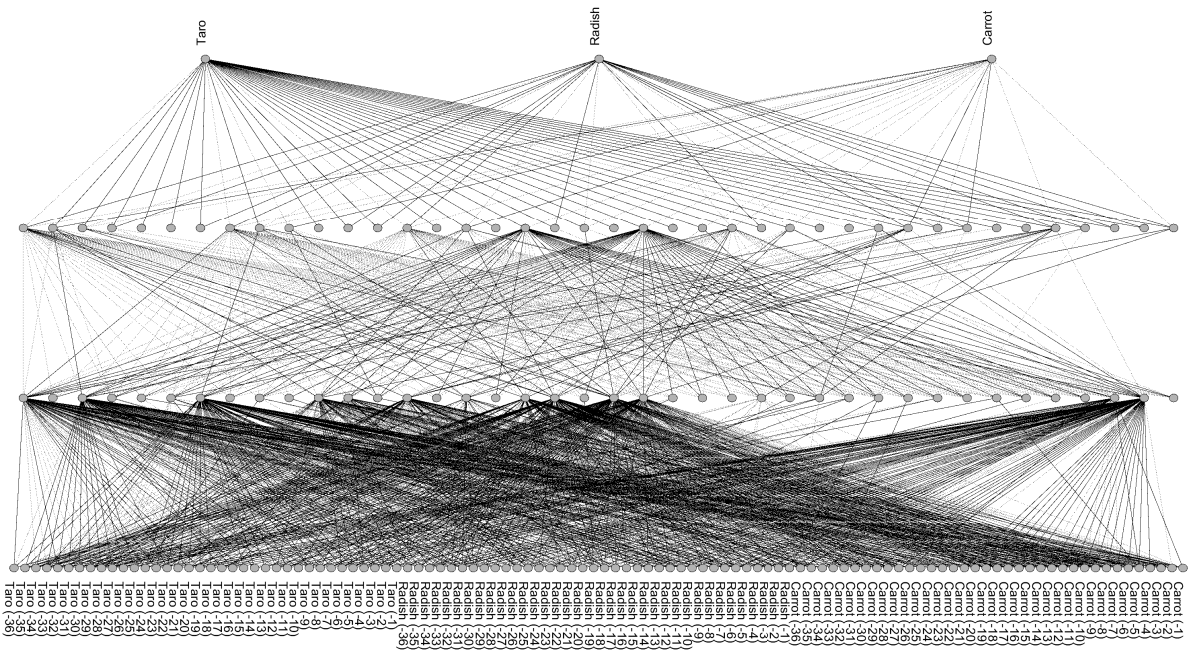}
  \caption{A layered neural network trained with a consumer price index data set. 
  This neural network was trained to predict the consumer price indices of taro, radish, and carrot in a month, from the input data of the previous $36$ months of 
  their consumer price indices. 
  The label written near the input layer ``Food name (-$n$)" shows the consumer price index of the food $n$ months before. }
  \label{fig:lnn_food}
\end{figure}

\begin{figure}[t]
  \hspace{-15mm}\includegraphics[width=160mm]{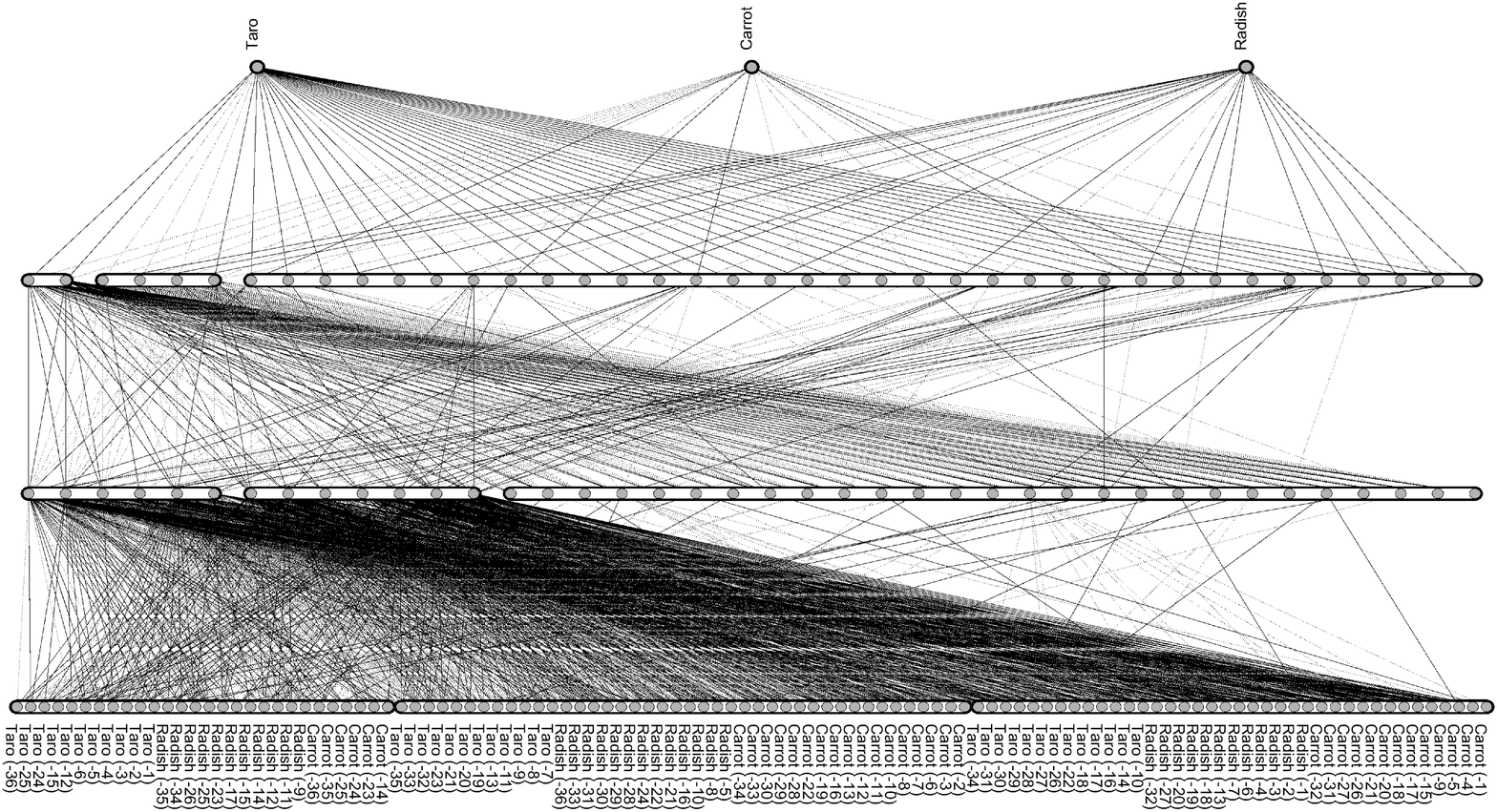}
  \caption{Community structure extracted from the trained layered neural network. 
  In each layer, the units were decomposed into three communities. }
  \label{fig:community_food}
\end{figure}

\begin{figure}[t]
  \hspace{-15mm}\includegraphics[width=160mm]{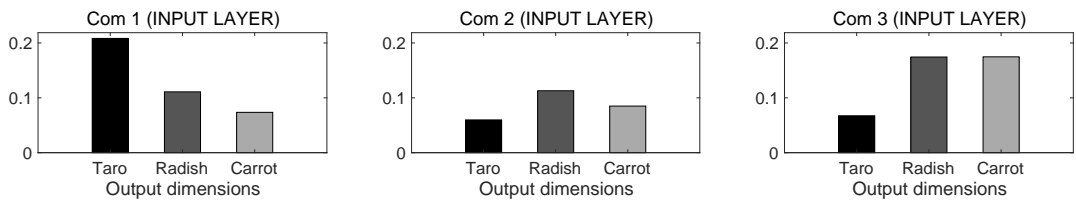}
  \caption{Effect of each community in the input layer on the output dimensions. 
    Com $1$, $2$, and $3$ represent the communities on the left, middle, and right in Figure \ref{fig:community_food}, respectively. 
    This figure shows that Com $1$ and $2$ are mainly used for predicting the consumer price indices of taro and radish, respectively. 
    Com $3$ is used for the prediction of both radish and carrot. }
  \label{fig:role_food1}
\end{figure}

\begin{figure}[t]
  \hspace{-15mm}\includegraphics[width=160mm]{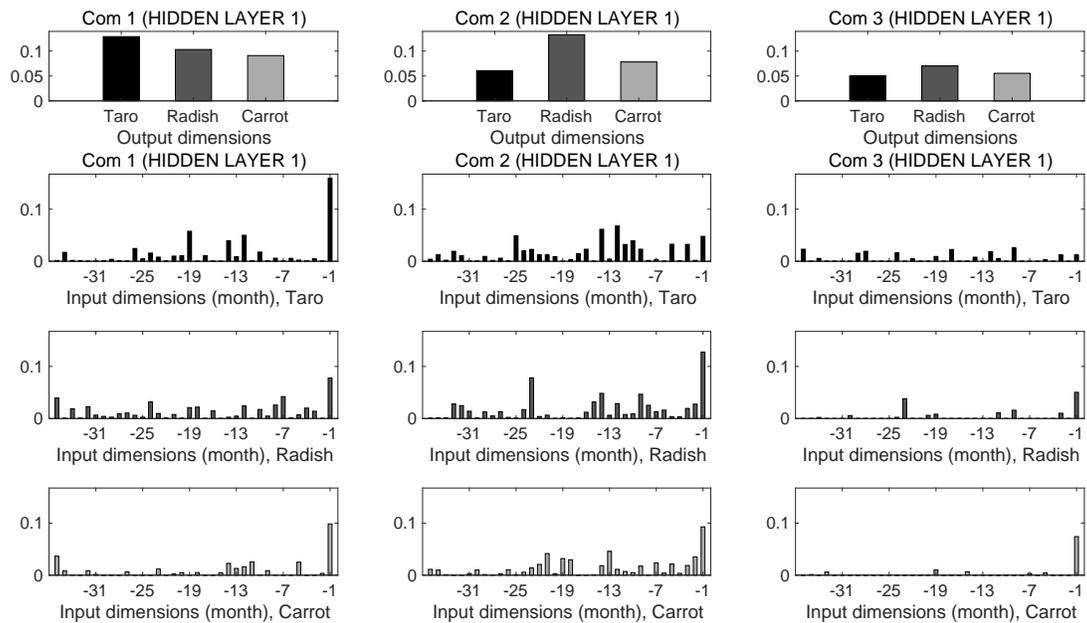}
  \caption{Top: Effect of each community in hidden layer $1$ (or the hidden layer adjacent to the input layer) on the output dimensions. 
  Second and third from the top, and the bottom: Effect of the input dimensions that correspond to the consumer price indices of taro, radish, and 
  carrot, respectively, on each community. 
  The labels ``$-n$" on the horizontal axis show the data $n$ months previously. 
  For example, Com $2$ is mainly used for predicting the consumer price index of radish, by using the input information about the consumer price indices of 
  radish one and $23$ months before, and that of carrot one month before. }
  \label{fig:role_food2}
\end{figure}

\begin{figure}[t]
  \hspace{-15mm}\includegraphics[width=160mm]{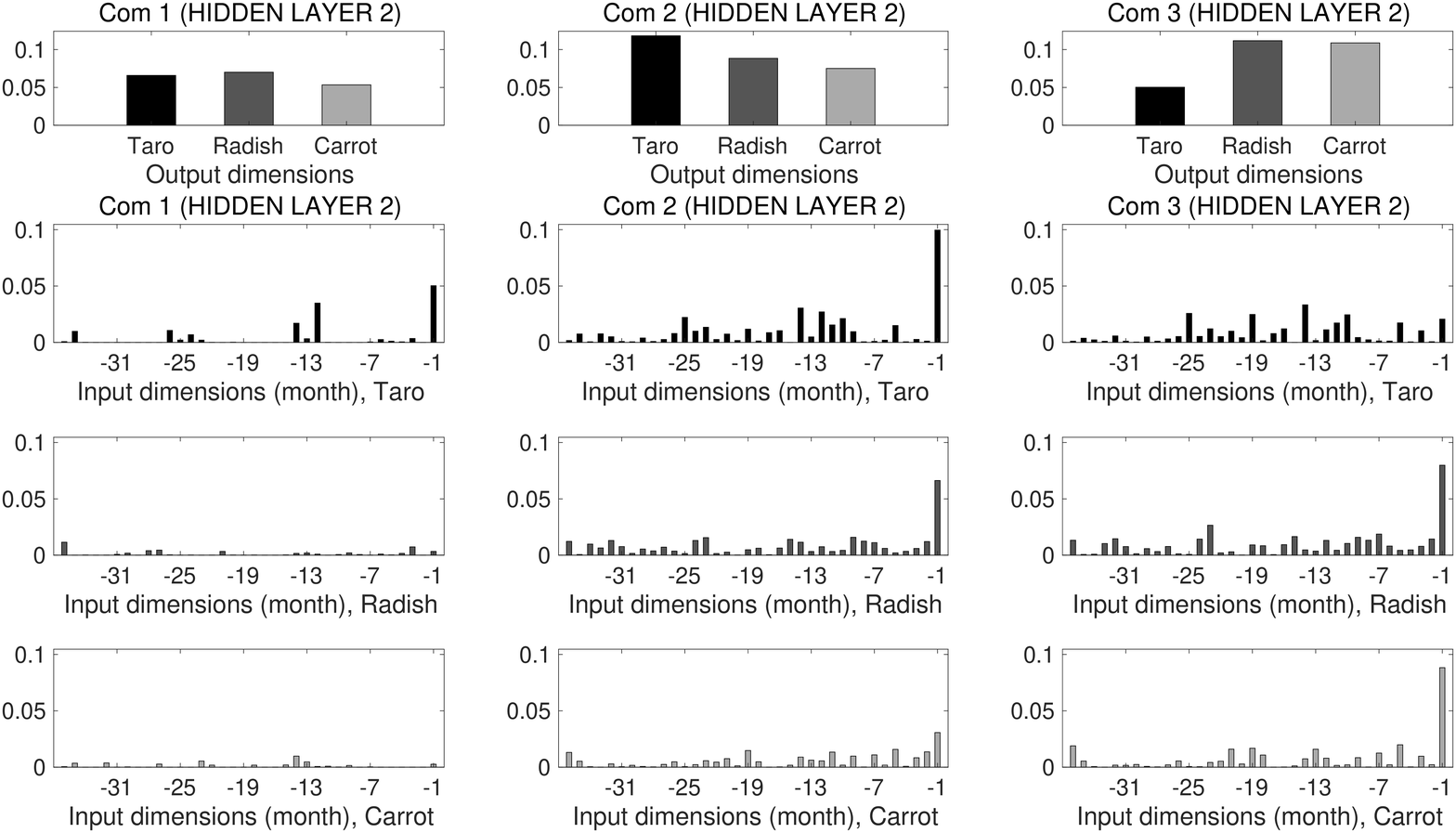}
  \caption{Top: Effect of each community in hidden layer $2$ (or the hidden layer adjacent to the output layer) on the output dimensions. 
  Second and third from the top, and the bottom: Effect of the input dimensions on each community in hidden layer $2$. }
  \label{fig:role_food3}
\end{figure}

\begin{figure}[t]
  \hspace{-15mm}\includegraphics[width=160mm]{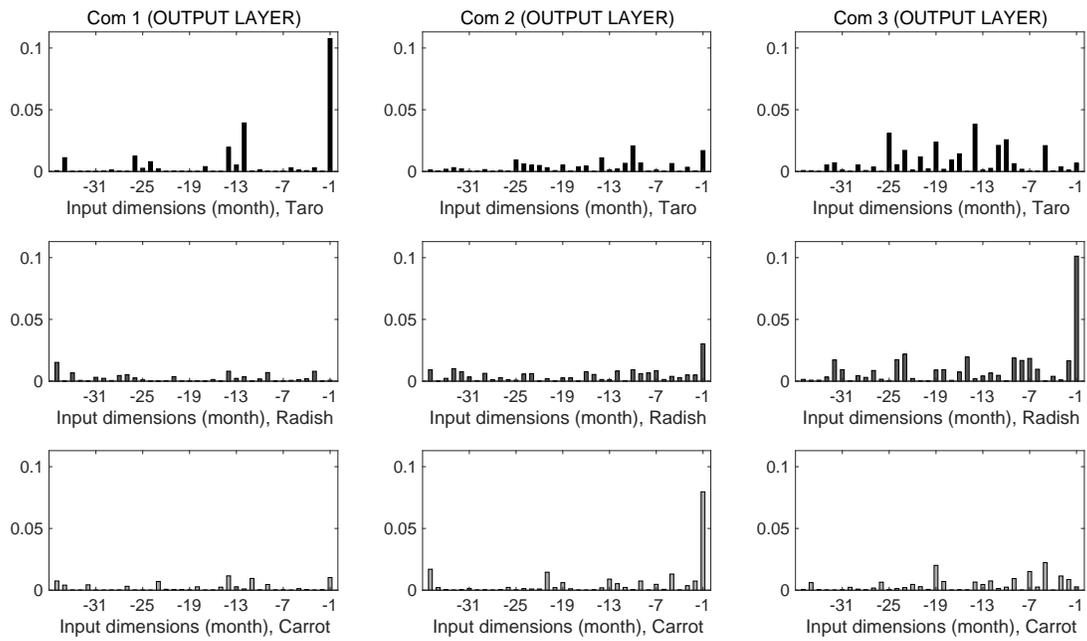}
  \caption{Effect of the input dimensions on each community in the output layer. 
  Com $1$, $2$, and $3$ correspond to the predicted consumer price indices of taro, carrot, and radish, respectively. 
  This figure shows that, to predict the consumer price index of each kind of food, its consumer price index one month before is the most commonly used 
  input information. }
  \label{fig:role_food4}
\end{figure}

Next, we applied our proposed method to a layered neural network trained with a sequential data set that shows the transition of consumer price indices of foods (\cite{estat}). 
The trained neural network predicts consumer price indices of taro, Japanese radish (in this paper, referred to as radish, for simplicity) and carrot for a month from 
$36$ months' input data. 
Figure \ref{fig:data_food} shows the output data or ground truth consumer price indices for $541$ months. 
With this data set, we trained a layered neural network, detected communities in a trained layered neural network, and quantified the relationship between each 
community and the input or output dimensions. 
Figures \ref{fig:lnn_food} and \ref{fig:community_food} show the trained layered neural network and the extracted community structure. 
Figures \ref{fig:role_food1}, \ref{fig:role_food2}, \ref{fig:role_food3}, and \ref{fig:role_food4} show the role of each community in the input, hidden and output 
layers, respectively. 

Figures \ref{fig:role_food1} to \ref{fig:role_food4} show that each community in the output layer mainly uses the input information of the corresponding 
class (or kind of food) from one month before, to infer the consumer price index of that class. 
By observing the role of Com $1$ in the output layer, it can be seen that the information for a year before is also used to infer information about taro. 
The result of Com $3$ in the output layer shows that the layered neural network uses information about other kinds of food (taro), to predict information about radish. 
In hidden layer $2$ (or the hidden layer that is adjacent to the output layer), Com $2$ and $3$ are mainly used to infer information about taro, and other 
vegetables, respectively. In hidden layer $1$ (or the hidden layer that is adjacent to the input layer), Com $1$ extracts information about the consumer price indices 
of all vegetables one month ago, while Com $2$ also uses information from longer ago, and in particular data from one and two years ago. 
The input layer results show the consumer price indices of the output classes that were inferred from the information about each input community. 
From Figure \ref{fig:role_food4}, Com $1$ and $3$ are used to infer information about taro and other vegetables. 


\subsection{Experiment Using Image Data Set}
\label{sec:exp3}

\begin{figure}[t]
  \hspace{-15mm}\includegraphics[width=160mm]{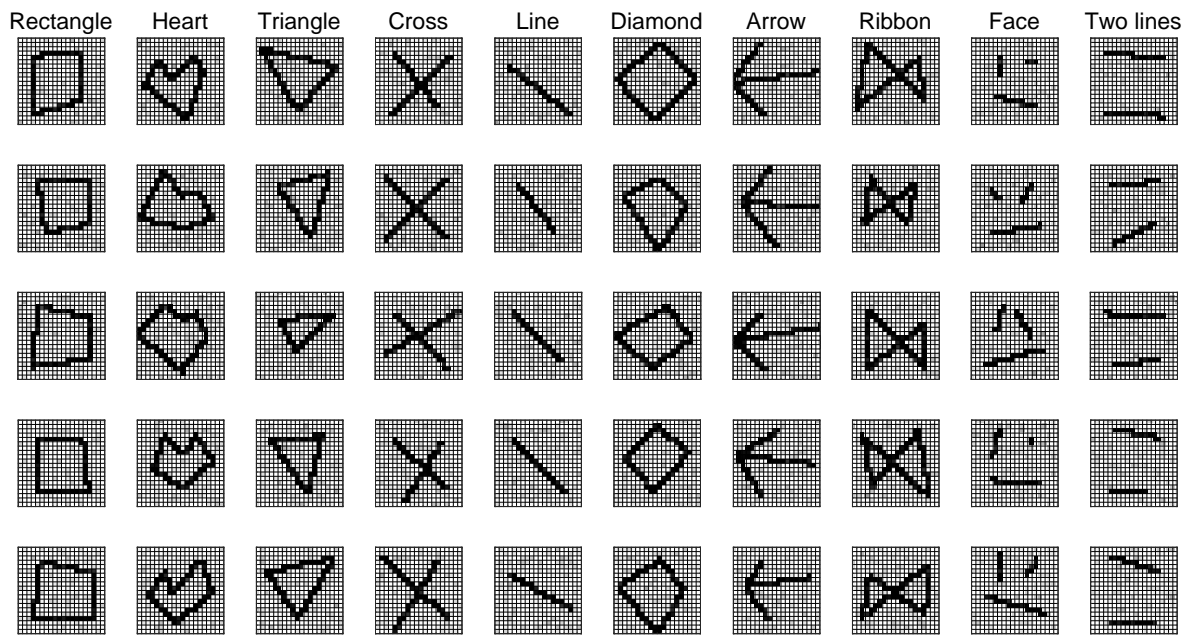}
  \caption{Sample input image data for each class.}\vspace{2mm}
  \label{fig:data_sample_diagram}
\end{figure}

We also applied our proposed method to an image data set of $10$ types of diagrams. 
This data set consists of images composed of $20\times 20$ pixels, and each image was generated by the following procedure. First, we generated a binary image by 
connecting randomly generated points from given distributions. 
Then, we added noise $\epsilon_3$ to each pixel, given by $\epsilon_3 \overset{\text\small\rm{i.i.d.}}{\sim}\mathcal{N}(0,0.1)$. 
Here, we determined the mean $x$ and $y$ coordinates and connected pairs of all points for each diagram as follows. 
We defined the maximum and minimum values of the $x$ and $y$ coordinates for each image as $1$ and $0$, respectively. 
\begin{itemize}
\item For class $1$ (``Rectangle"), points are given by: $p_1=(0.2, 0.2)$, $p_2=(0.2, 0.8)$, $p_3=(0.8, 0.8)$, $p_4=(0.8, 0.2)$. 
  Pairs of mutually connected points: $(p_1, p_2)$, $(p_2, p_3)$, $(p_3, p_4)$, $(p_4, p_1)$. 
\item For class $2$ (``Heart"), points are given by: $p_1=(0.1, 0.5)$, $p_2=(0.3, 0.8)$, $p_3=(0.5, 0.6)$, $p_4=(0.7, 0.8)$, $p_5=(0.9, 0.5)$, $p_6=(0.5, 0.2)$. 
  Pairs of mutually connected points: $(p_1, p_2)$, $(p_2, p_3)$, $(p_3, p_4)$, $(p_4, p_5)$ $(p_5, p_6)$, $(p_6, p_1)$. 
\item For class $3$ (``Triangle"), points are given by: $p_1=(0.5, 0.2)$, $p_2=(0.8, 0.8)$, $p_3=(0.2, 0.8)$. 
  Pairs of mutually connected points: $(p_1, p_2)$, $(p_2, p_3)$, $(p_3, p_1)$. 
\item For class $4$ (``Cross"), points are given by: $p_1=(0.2, 0.2)$, $p_2=(0.8, 0.8)$, $p_3=(0.2, 0.8)$, $p_4=(0.8, 0.2)$. 
  Pairs of mutually connected points: $(p_1, p_2)$, $(p_3, p_4)$. 
\item For class $5$ (``Line"), points are given by: $p_1=(0.2, 0.8)$, $p_2=(0.8, 0.2)$. 
  Pairs of mutually connected points: $(p_1, p_2)$. 
\item For class $6$ (``Diamond"), points are given by: $p_1=(0.5, 0.9)$, $p_2=(0.9, 0.5)$, $p_3=(0.5, 0.1)$, $p_4=(0.1, 0.5)$. 
  Pairs of mutually connected points: $(p_1, p_2)$, $(p_2, p_3)$, $(p_3, p_4)$, $(p_4, p_1)$. 
\item For class $7$ (``Arrow"), points are given by: $p_1=(0.4, 0.9)$, $p_2=(0.1, 0.5)$, $p_3=(0.4, 0.1)$, $p_4=(0.9, 0.5)$. 
  Pairs of mutually connected points: $(p_1, p_2)$, $(p_2, p_3)$, $(p_2, p_4)$. 
\item For class $8$ (``Ribbon"), points are given by: $p_1=(0.2, 0.2)$, $p_2=(0.8, 0.8)$, $p_3=(0.8, 0.2)$, $p_4=(0.2, 0.8)$. 
  Pairs of mutually connected points: $(p_1, p_2)$, $(p_2, p_3)$, $(p_3, p_4)$, $(p_4, p_1)$. 
\item For class $9$ (``Face"), points are given by: $p_1=(0.3, 0.8)$, $p_2=(0.3, 0.6)$, $p_3=(0.7, 0.8)$, $p_4=(0.7, 0.6)$, $p_5=(0.2, 0.3)$, $p_6=(0.8, 0.3)$. 
  Pairs of mutually connected points: $(p_1, p_2)$, $(p_3, p_4)$, $(p_5, p_6)$. 
\item For class $10$ (``Two lines"), points are given by: $p_1=(0.2, 0.2)$, $p_2=(0.8, 0.2)$, $p_3=(0.2, 0.8)$, $p_4=(0.8, 0.8)$. 
  Pairs of mutually connected points: $(p_1, p_2)$, $(p_3, p_4)$. 
\end{itemize}
For all images, the points were independently generated from a normal distribution with the above mean and a standard deviation of $0.07$. 
Figure \ref{fig:data_sample_diagram} shows sample images for each class of diagrams. 

\begin{figure}[t]
  \hspace{-15mm}\includegraphics[width=160mm]{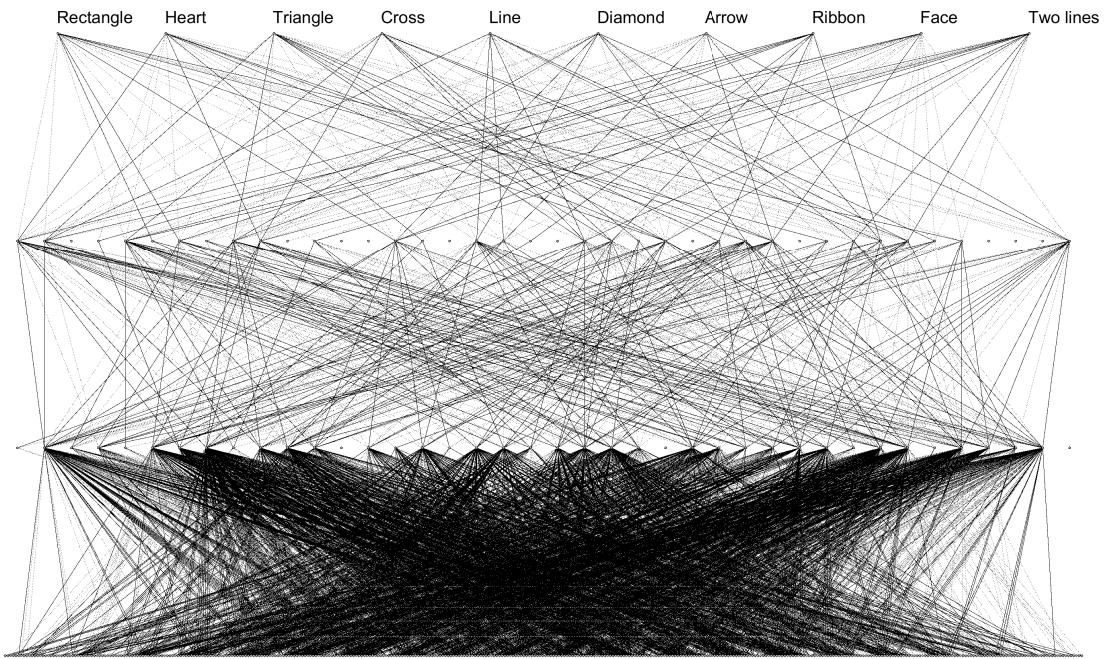}
  \caption{A layered neural network trained with an image data set. 
  This neural network was trained to classify the input image to $10$ classes of diagrams. 
  The input dimensions represent the pixel values of an input image. }\vspace{3mm}
  \label{fig:lnn_diagram}
\end{figure}

\begin{figure}[t]
  \hspace{-15mm}\includegraphics[width=160mm]{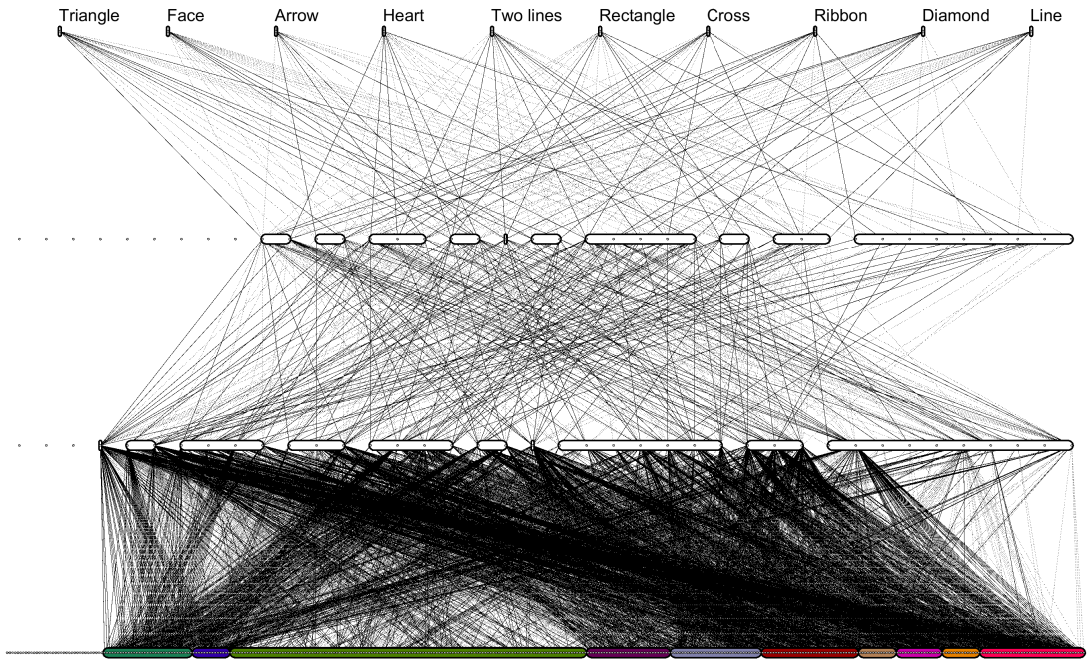}
  \caption{Community structure extracted from the trained layered neural network (best viewed in color).
  In each layer, the units were decomposed into $10$ communities. 
  The colors of each community in the input layer in this figure correspond to those of the pixels in Figure \ref{fig:com_pixel_diagram}. }
  \label{fig:community_diagram}
\end{figure}

\begin{figure}[t]
  \hspace{-15mm}\includegraphics[width=160mm]{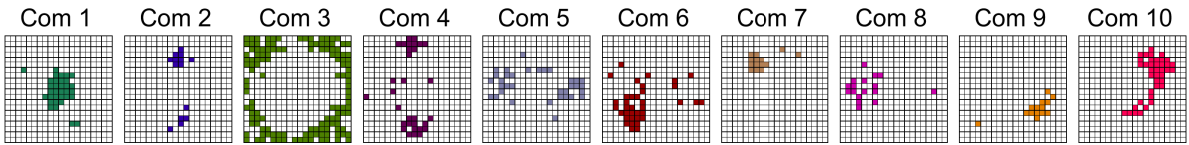}
  \caption{Communities of pixels in the input layer (best viewed in color). 
  The colors of the pixels in this figure correspond to those of each community in the input layer in Figure \ref{fig:community_diagram}. }
  \label{fig:com_pixel_diagram}
\end{figure}

\begin{figure}[t]
  \hspace{-15mm}\includegraphics[width=160mm]{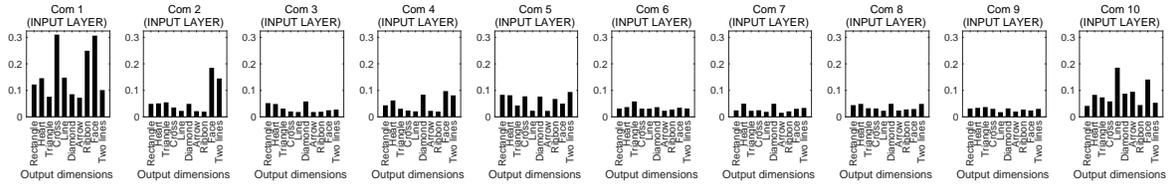}
  \caption{Effect of each community in the input layer on the output dimensions. 
    Com $1$, $\cdots,$ $10$ represent the communities that are drawn from left to right in Figure \ref{fig:com_pixel_diagram}. 
    For example, Com $1$ is used to classify multiple diagrams, especially ``Cross," ``Ribbon," and ``Face." }
  \label{fig:role_diagram1}
\end{figure}

\begin{figure}[t]
  \hspace{-15mm}\includegraphics[width=160mm]{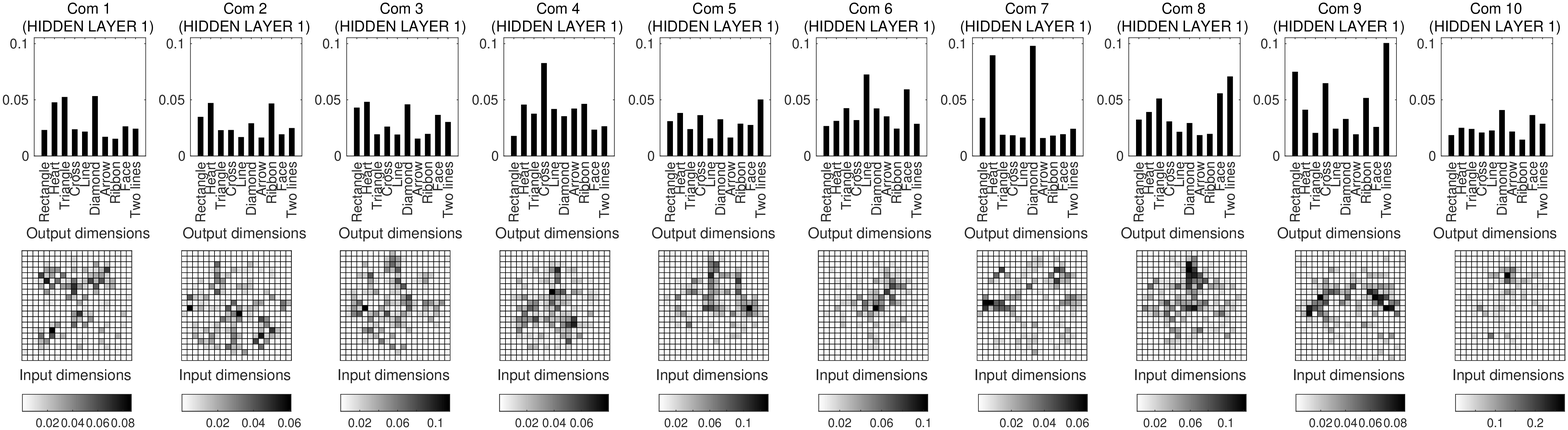}
  \caption{Top: Effect of each community in hidden layer $1$ (or the hidden layer adjacent to the input layer) on the output dimensions. 
  Bottom: Effect of the input dimensions on each community in hidden layer $1$. }
  \label{fig:role_diagram2}
\end{figure}

\begin{figure}[t]
  \hspace{-15mm}\includegraphics[width=160mm]{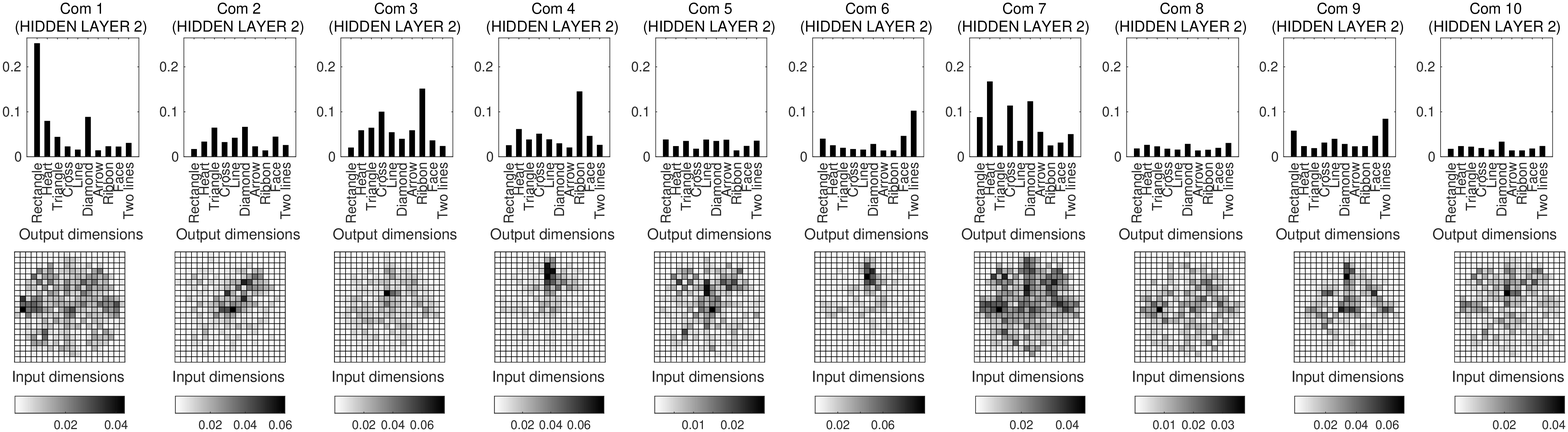}
  \caption{Top: Effect of each community in hidden layer $2$ (or the hidden layer adjacent to the output layer) on the output dimensions. 
  Bottom: Effect of the input dimensions on each community in hidden layer $2$. }
  \label{fig:role_diagram3}
\end{figure}

\begin{figure}[t]
  \hspace{-15mm}\includegraphics[width=160mm]{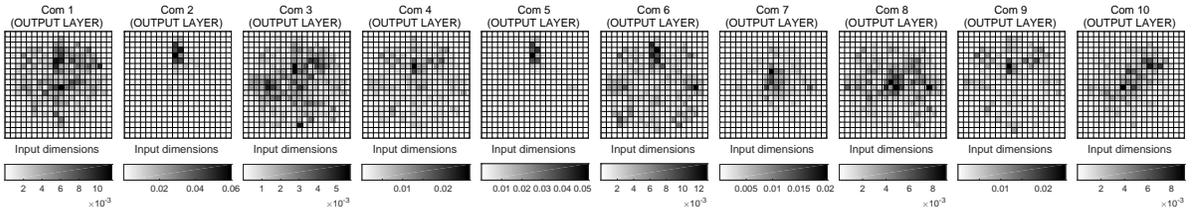}
  \caption{Effect of the input dimensions on each community in the output layer. 
  Com $1$, $\cdots,$ $10$ correspond to the classification results for ``Triangle," ``Face," ``Arrow," ``Heart," ``Two lines," ``Rectangle," ``Cross," ``Ribbon," 
  ``Diamond," and ``Line," respectively. }
  \label{fig:role_diagram4}
\end{figure}

With this data set, we trained a layered neural network, detected communities in a trained layered neural network, and quantified the relationship between each community and 
the input or output dimensions. 
Figures \ref{fig:lnn_diagram} and \ref{fig:community_diagram} show the trained layered neural network and the extracted community structure, respectively. 
Figures \ref{fig:role_diagram1}, \ref{fig:role_diagram2}, \ref{fig:role_diagram3}, and \ref{fig:role_diagram4} show the role of each community in the input, 
hidden and output layers, respectively. 
The community assignment of each pixel of an input image is shown in Figure \ref{fig:com_pixel_diagram}. 

From Figures \ref{fig:role_diagram1} to \ref{fig:role_diagram4}, we can learn the following about the inference:
\begin{itemize}
\item From the figures of the input layer, 
  \begin{itemize}
  \item Information about pixels in Com $3$, $6$, $7$, $8$, and $9$ are relatively little used for inferring any output dimension values. 
  Among these communities, Com $3$ consists of the pixels at the periphery of an image, as shown in Figure \ref{fig:com_pixel_diagram}. 
  On the other hand, Com $1$ is used for inferring multiple output dimensions (especially, ``Cross," ``Ribbon," and ``Face"), and it consists of the pixels 
  at the center of the image. 
  Here, among the mean images of these three classes of diagrams, ``Cross" and ``Ribbon" include an X-shaped partial image at the center point, 
  while ``Face" has uncolored pixels at the same point. 
  Note that our proposed method reveals which pixels of an input image are used in combination to infer 
  something, but we do not know how these pixels were used. As with Com $1$, part of the trained layered neural network may use the information that there are some patterns in 
  these pixels, or it may use the information that there are no colored pixels in this region. 
  \item Pixels in Com $2$ are used mainly to infer ``Face" and ``Two lines", and they are located in the middle in the horizontal direction, and at the top and 
  bottom edges in the vertical direction. From the settings of the distribution of the points in input data images, it can be inferred (though it is not directly 
  shown by the proposed method) that if the upper region is colored and the lower region is uncolored, then the layered neural network might predict that 
  the image is classified as ``Face." On the other hand, the layered neural network might predict that the image class is ``Two lines," if both the upper and 
  lower regions are colored. To check whether or not these hypotheses are correct, we need another analytical method to provide information about 
  how the values of the input dimensions are used in combination for the inference. 
\end{itemize}
\item From the figures of hidden layer $1$, 
  \begin{itemize}
  \item Each community captures a partial area in an image: For example, Com $7$ uses the partial area of an input image that is located near the mean points of the 
  upper part of ``Heart," and it is used for the inference of ``Heart" and ``Diamond." 
  \item Com $9$ uses the information about the region that is located near the center in the vertical direction, and it mainly consists of two small diagonal lines 
  near the edges of an image. 
  This community is used mainly for the classification of ``Rectangle," ``Cross," ``Ribbon," and ``Two lines." Among these diagrams, the mean images of ``Rectangle" and ``Ribbon" 
  have colored pixels in this region, while ``Two lines" has uncolored pixels. 
  \end{itemize}
\item From the figures of hidden layer $2$, 
  \begin{itemize}
  \item ``Rectangle" is mainly classified by Com $1$, which captures global information from an input image. Com $7$ also uses the information obtained from pixels 
  in a region with a wide range to classify the diagrams of ``Rectangle," ``Heart," ``Cross," and ``Diamond." 
  \item Com $4$ is used to classify an image as ``Ribbon," by observing the image region that is located near the center in the middle in the horizontal direction, 
  and at the top in the vertical direction. In the mean image of ``Ribbon," this region is uncolored, therefore, it is inferred that the neural network classifies an 
  image as ``Ribbon," if there are uncolored pixels in this region. 
  \end{itemize}
\item From the output layer figures, 
  \begin{itemize}
  \item We can determine the part of an input image from which the layered neural network used information and thus classify the images to each class of diagram. 
  As shown in Figure \ref{fig:community_diagram}, Com $1$, $\cdots$, $10$ respectively correspond to the ``Triangle," ``Face," ``Arrow," ``Heart," ``Two lines," 
  ``Rectangle," ``Cross," ``Ribbon," ``Diamond," and ``Line" classes. For instance, both Com $2$ (``Face") and $5$ (``Two lines") use the information from an image region that is 
  located in the middle in the horizontal direction, and at the top in the vertical direction. The mean image of ``Face" has uncolored pixels in this region, while 
  that of ``Two lines" has colored pixels. These results show that the colors of the pixels in this region contribute to the classification of these diagrams. 
  \item To classify the class represented by Com $7$ (``Cross"), the region at the center of an image is used. It can be inferred that if there is an X-shape 
  in this region, then the layered neural network classifies the image as ``Cross." 
  \end{itemize}
\end{itemize}


\section{Discussion}
\label{sec:discussions}

In this section, we discuss the limitations and future work in relation to the proposed method from the standpoints of analytical method, interpretability, and applications. 

First, in this paper, we proposed analyzing the role of each community extracted by a network analysis method. 
Specifically, in the experiment described in section \ref{sec:exp3}, there were several communities in a layer that use similar input information, or contribute to similar 
sets of output dimensions, which made it difficult to interpret the specific role of each community in the trained network. 
Natural extensions of the proposed method are first to quantify the role of each \textbf{unit} by assuming that each unit composes one community, and then to classify 
the units into clusters, based on the similarities between their feature vectors. 
With these extensions, we can obtain clusters of units that have relatively similar relationships with the input and output dimensions, therefore the role of each 
community might be more clearly differentiated. 

Second, our proposed method has enabled us to interpret the role of each community as regards the relationships with the input and output dimensions, 
however, this does not mean that it can provide all knowledge about the extracted communities. Specifically, the proposed method cannot tell us in detail 
about how input information is processed from one community to another. 
We need another method to explain the sequential meachanism of the inter-layer functions of a trained neural network. 
Additionally, for the communities in the output layer, our proposed method provides us with the information that is somewhat similar to that of linear models. 
However, it is important to construct a method to interpret the internal mechanism of a layered neural network, since a linear model cannot express the complex input-output 
relationships in the high-dimensional practical data sets (Appendix. C). 

Finally, it might be possible to improve the generalization performance of a neural network based on the analysis result obtained with the proposed method. 
For instance, the information about the role of each community can be used for determining hyperparameters or the number of communities in community detection: 
if there are multiple communities in a layer that play similar roles in inference, they might be redundant and be more appropriately represented by fewer communities. 


\section{Conclusion}
\label{sec:conclusion}

Layered neural networks have contributed to a great improvement in the prediction of various practical data sets by their powerful ability to express high-dimensional complex data. 
However, its application area has thus for been limited, since it is difficult for human beings to understand the internal inference mechanism. 
Our previous methods have enabled us to obtain a simplified network structure for a trained layered neural network based on network analysis. 
However, these methods could not provide us with quantitative information about the role of each part in a neural network. 
In this paper, we proposed an analytical method for interpreting the role of each community, where the communities are extracted by our previous methods. 
It enables us to gain knowledge about the function of each community by analyzing the contribution of each input dimension to a community and also about the contribution of 
a community to each output dimension. 
We experimentally showed that our proposed method provided an interpretation of the role of each community by using both synthetic and practical data sets. 


\newpage
\appendix
\section*{Appendix A. Proof of EM Algorithm for Community Detection}
\label{app:proof}

\begin{proof}
The denominator and numerator in the last term of Equation (\ref{eq:qdefine}) are given, respectively, by
\begin{eqnarray*}
  \lefteqn{ \mathrm{Pr}(A^+,A^-,B^+,B^-,g_k=c|\pi,\tau^+,\tau^-,\tau'^+,\tau'^-) }\\
  &=& \sum_{g_1}\cdots \sum_{g_{k_0}} \delta_{g_k,c} \mathrm{Pr}(A^+,A^-,B^+,B^-,g|\pi,\tau^+,\tau^-,\tau'^+,\tau'^-) \\
  &=& \sum_{g_1}\cdots \sum_{g_{k_0}} \delta_{g_k,c} \prod_h \left\{ \pi_{g_h} 
  \left[ \prod_i \Bigl(\tau^+_{g_h,i}\Bigr)^{A^+_{i,h}} \Bigl(1-\tau^+_{g_h,i}\Bigr)^{1-A^+_{i,h}} 
  \Bigl(\tau^-_{g_h,i}\Bigr)^{A^-_{i,h}} \Bigl(1-\tau^-_{g_h,i}\Bigr)^{1-A^-_{i,h}} \right] \right. \\
  &&\left. \left[ \prod_j \Bigl(\tau'^+_{g_h,j}\Bigr)^{B^+_{h,j}} \Bigl(1-\tau'^+_{g_h,j}\Bigr)^{1-B^+_{h,j}} 
  \Bigl(\tau'^-_{g_h,j}\Bigr)^{B^-_{h,j}} \Bigl(1-\tau'^-_{g_h,j}\Bigr)^{1-B^-_{h,j}} \right] \right\}\\
  &=& \left\{ \pi_c \left[ \prod_i \Bigl(\tau^+_{c,i}\Bigr)^{A^+_{i,k}} \Bigl(1-\tau^+_{c,i}\Bigr)^{1-A^+_{i,k}} 
  \Bigl(\tau^-_{c,i}\Bigr)^{A^-_{i,k}} \Bigl(1-\tau^-_{c,i}\Bigr)^{1-A^-_{i,k}} \right] \right. \\
  &&\left. \left[ \prod_j \Bigl(\tau'^+_{c,j}\Bigr)^{B^+_{k,j}} \Bigl(1-\tau'^+_{c,j}\Bigr)^{1-B^+_{k,j}} 
  \Bigl(\tau'^-_{c,j}\Bigr)^{B^-_{k,j}} \Bigl(1-\tau'^-_{c,j}\Bigr)^{1-B^-_{k,j}} \right] \right\} \\
  &&\left\{ \prod_{h\neq k} \sum_s \pi_s \left[ \prod_i \Bigl(\tau^+_{s,i}\Bigr)^{A^+_{i,h}} \Bigl(1-\tau^+_{s,i}\Bigr)^{1-A^+_{i,h}} 
  \Bigl(\tau^-_{s,i}\Bigr)^{A^-_{i,h}} \Bigl(1-\tau^-_{s,i}\Bigr)^{1-A^-_{i,h}} \right] \right. \\
  &&\left. \left[ \prod_j \Bigl(\tau'^+_{s,j}\Bigr)^{B^+_{h,j}} \Bigl(1-\tau'^+_{s,j}\Bigr)^{1-B^+_{h,j}} 
  \Bigl(\tau'^-_{s,j}\Bigr)^{B^-_{h,j}} \Bigl(1-\tau'^-_{s,j}\Bigr)^{1-B^-_{h,j}} \right] \right\},
\end{eqnarray*}
and
\begin{eqnarray*}
  \lefteqn{ \mathrm{Pr}(A^+,A^-,B^+,B^-|\pi,\tau^+,\tau^-,\tau'^+,\tau'^-) }\\
  &=& \sum_{g_1}\cdots \sum_{g_{k_0}} \mathrm{Pr}(A^+,A^-,B^+,B^-,g|\pi,\tau^+,\tau^-,\tau'^+,\tau'^-)\\
  &=& \prod_h \sum_s \pi_s \left[ \prod_i \Bigl(\tau^+_{s,i}\Bigr)^{A^+_{i,h}} \Bigl(1-\tau^+_{s,i}\Bigr)^{1-A^+_{i,h}} 
  \Bigl(\tau^-_{s,i}\Bigr)^{A^-_{i,h}} \Bigl(1-\tau^-_{s,i}\Bigr)^{1-A^-_{i,h}} \right] \\
  &&\left[ \prod_j \Bigl(\tau'^+_{s,j}\Bigr)^{B^+_{h,j}} \Bigl(1-\tau'^+_{s,j}\Bigr)^{1-B^+_{h,j}} 
  \Bigl(\tau'^-_{s,j}\Bigr)^{B^-_{h,j}} \Bigl(1-\tau'^-_{s,j}\Bigr)^{1-B^-_{h,j}} \right],
\end{eqnarray*}
where $\delta_{i,j}$ is the Kronecker delta. Therefore, $q_{k,c}$ is given by Equation (\ref{eq:q}).

The problem is to maximize $\bar{\mathcal{L}}$ of Equation (\ref{eq:exploglh}) with a given $\{q_{k,c}\}$ under the condition of Equation (\ref{eq:normalization}). 
This is solved with the Lagrangian undetermined multiplier method, which employs
\begin{eqnarray*}
  f = \bar{\mathcal{L}}-\alpha \sum_c \pi_c,
\end{eqnarray*}
and
\begin{eqnarray}
  \frac{\partial f}{\partial \pi_c}=\frac{\partial f}{\partial \tau^+_{c,i}}=\frac{\partial f}{\partial \tau^-_{c,i}}
  =\frac{\partial f}{\partial \tau'^+_{c,j}}=\frac{\partial f}{\partial \tau'^-_{c,j}}=0.
  \label{eq:lag1}
\end{eqnarray}
From Equation (\ref{eq:lag1}), the following equations are derived:
\begin{eqnarray}
\frac{\partial \bar{\mathcal{L}}}{\partial \pi_c}=\alpha,\ \ \ 
\frac{\partial \bar{\mathcal{L}}}{\partial \tau^+_{c,i}}=\frac{\partial \bar{\mathcal{L}}}{\partial \tau^-_{c,i}}=
\frac{\partial \bar{\mathcal{L}}}{\partial \tau'^+_{c,j}}=\frac{\partial \bar{\mathcal{L}}}{\partial \tau'^-_{c,j}}=0.
  \label{eq:lag2}
\end{eqnarray}
Using Equations (\ref{eq:exploglh}) and (\ref{eq:lag2}), we obtain 
\begin{eqnarray}
\lefteqn{ \pi_c = \frac{1}{\alpha}\sum_k q_{k,c},\ \ 
\tau^+_{c,i} = \frac{\sum_k A^+_{i,k} q_{k,c}}{\sum_k q_{k,c}},\ \ 
\tau^-_{c,i} = \frac{\sum_k A^-_{i,k} q_{k,c}}{\sum_k q_{k,c}}, }\nonumber \\
&&\tau'^+_{c,j} = \frac{\sum_k B^+_{k,j} q_{k,c}}{\sum_k q_{k,c}},\ \ 
\tau'^-_{c,j} = \frac{\sum_k B^-_{k,j} q_{k,c}}{\sum_k q_{k,c}}.
  \label{eq:lag3}
\end{eqnarray}
From Equation (\ref{eq:lag3}) and the condition of Equation (\ref{eq:normalization}), Lagrange's undetermined multiplier $\alpha$ is determined, 
and Equation (\ref{eq:lag3}) is rewritten as Equation (\ref{eq:pitau}). 
\end{proof}


\section*{Appendix B. Experimental Settings}
\label{app:experimentalsettings}

Table \ref{tab:settings} shows the detailed settings for the experiments described in sections \ref{sec:exp1}, \ref{sec:exp2}, and \ref{sec:exp3}. 
In Table \ref{tab:settings}, we denote the experiments in sections \ref{sec:exp1}, \ref{sec:exp2}, and \ref{sec:exp3} as Exp.1, 2, and 3, respectively. 

We perform the following process common to the all experiments. 
\begin{itemize}
\item We normalize the input data set so that the maximum and minimum values of an element are $x_{\mathrm{max}}$ and $x_{\mathrm{min}}$. 
In the same way, we normalize the output data set by using the maximum and minimum values $y_{\mathrm{max}}$ and $y_{\mathrm{min}}$. 
\item We generated the initial parameters of a layered neural network as follows: 
\begin{eqnarray*}
  &&\omega^d_{ij}\overset{\text\small\rm{i.i.d.}}{\sim}\mathcal{N}(0,0.5), \\
  &&\theta^d_i\overset{\text\small\rm{i.i.d.}}{\sim}\mathcal{N}(0,0.5).
\end{eqnarray*}
\item We substituted the elements of $1$ or $0$ of the adjacency matrices $A^+,A^-,B^+,B^-$ with $0.99$ and $0.01$ to stabilize the EM algorithm. 
\item We set the iteration number of the EM algorithm at $a_2$. 
We defined the number of community detection trials as $a_3$, and used the result with the maximum expected log likelihood $\bar{\mathcal{L}}$ at the final iteration. 
\item We draw the positive and negative connection weights with solid lines and dotted lines, respectively. 
\end{itemize}

Let $X^{(k)}_n$ and $Y^{(k)}_n$, respectively, be the $n$-th samples of input and output training data in class $k$. 
To stabilize the layered neural network training in Exp.2 described in section \ref{sec:exp2}, we did not choose training data randomly in each iteration. 
Instead, we chose the training data in the following order: 
\begin{eqnarray*}
\{X^{(1)}_1, Y^{(1)}_1\}, \{X^{(2)}_1, Y^{(2)}_1\}, &\cdots&, \{X^{(10)}_1, Y^{(10)}_1\}, \\
\{X^{(1)}_2, Y^{(1)}_2\}, \{X^{(2)}_2, Y^{(2)}_2\}, &\cdots&, \{X^{(10)}_2, Y^{(10)}_2\}, \\ 
&\vdots& \\
\{X^{(1)}_{n_1} Y^{(1)}_{n_1}\}, \{X^{(2)}_{n_1}, Y^{(2)}_{n_1}\}, &\cdots&, \{X^{(10)}_{n_1}, Y^{(10)}_{n_1}\}.
\end{eqnarray*}
After the $10\times {n_1}$-th iteration, we return to the first sample and repeat the same process. 

\begin{table}[h]
\caption{The experimental settings of the parameters.}\vspace{2mm}
  \centering
  \scalebox{0.75}{
  \begin{tabular}{c|p{8.5cm}|c|c|c} \Hline
    name & meaning & Exp.1 & Exp.2 & Exp.3 \\ \hline \hline
    $n_1$ & number of training data sets & $5000$ & $1000$ for each class & $270$\\ \hline
    $x_{\mathrm{min}}$ & \parbox{8.5cm}{\strut{}minimum value of normalized input data\strut} & \multicolumn{3}{c}{$-1$}\\ \hline
    $x_{\mathrm{max}}$ & \parbox{8.5cm}{\strut{}maximum value of normalized input data\strut} & \multicolumn{3}{c}{$1$}\\ \hline
    $y_{\mathrm{min}}$ & \parbox{8.5cm}{\strut{}minimum value of normalized output data\strut} & \multicolumn{3}{c}{$0.01$}\\ \hline
    $y_{\mathrm{max}}$ & \parbox{8.5cm}{\strut{}maximum value of normalized output data\strut} & \multicolumn{3}{c}{$0.99$}\\ \hline
    $a_1$ & \parbox{8.5cm}{\strut{}mean iteration number of layered neural network training per data set\strut} & $2000$ & $100$ & $500$ \\ \hline
    $\eta$ & step size of layered neural network training & \multicolumn{3}{c}{$0.7$}\\ \hline
    $\lambda$ & hyperparameter of LASSO & $9.0\times 10^{-7}$ & $1.1\times 10^{-5}$ & $4.0\times 10^{-5}$\\ \hline
    $\epsilon_1$ & \parbox{8.5cm}{\strut{}hyperparameter for convergence of neural network\strut} & \multicolumn{3}{c}{$0.001$}\\ \hline
    $\xi$ & weight removing hyperparameter & $0.3$ & $5.0\times 10^{-3}$ & $5.0\times 10^{-4}$\\ \hline
    $C$ & \parbox{8.5cm}{\strut{}number of communities per layer\strut} & $3$ & $10$ & $3$\\ \hline
    $a_2$ & \parbox{8.5cm}{\strut{}iteration number of EM algorithm\strut} & \multicolumn{3}{c}{$200$}\\ \hline
    $a_3$ & \parbox{8.5cm}{\strut{}number of community detection trials\strut} & \multicolumn{3}{c}{$300$}\\ \Hline
  \end{tabular}
  }
\label{tab:settings}
\end{table}


\section*{Appendix C. Comparison with Linear Model}
\label{app:linear}

A linear model is one of the examples of interpretable models, and it provides us with the information about the degree to which the prediction result is affected 
by the value of each input dimension, in terms of the prediction coefficients. 
However, it is difficult for a linear model to represent the high-dimensional practical data sets as well as a layered neural network. 

\begin{figure}[t]
  \hspace{-15mm}\includegraphics[width=160mm]{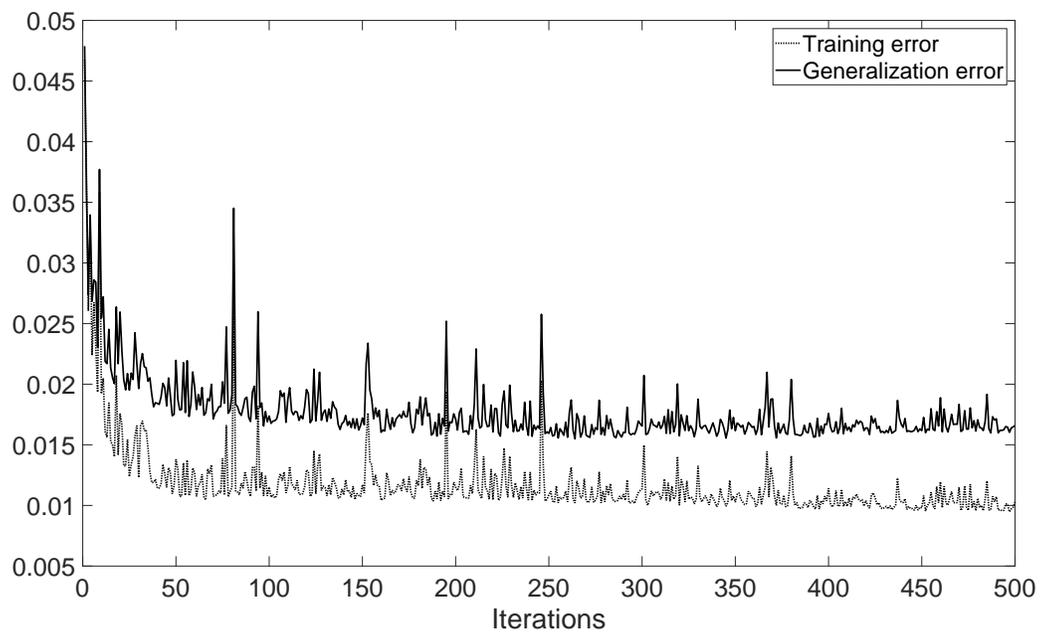}
  \caption{Training error and generalization error of a layered neural network. }
  \label{fig:appendixC_GT_LNN}
\end{figure}

\begin{figure}[t]
  \hspace{-15mm}\includegraphics[width=160mm]{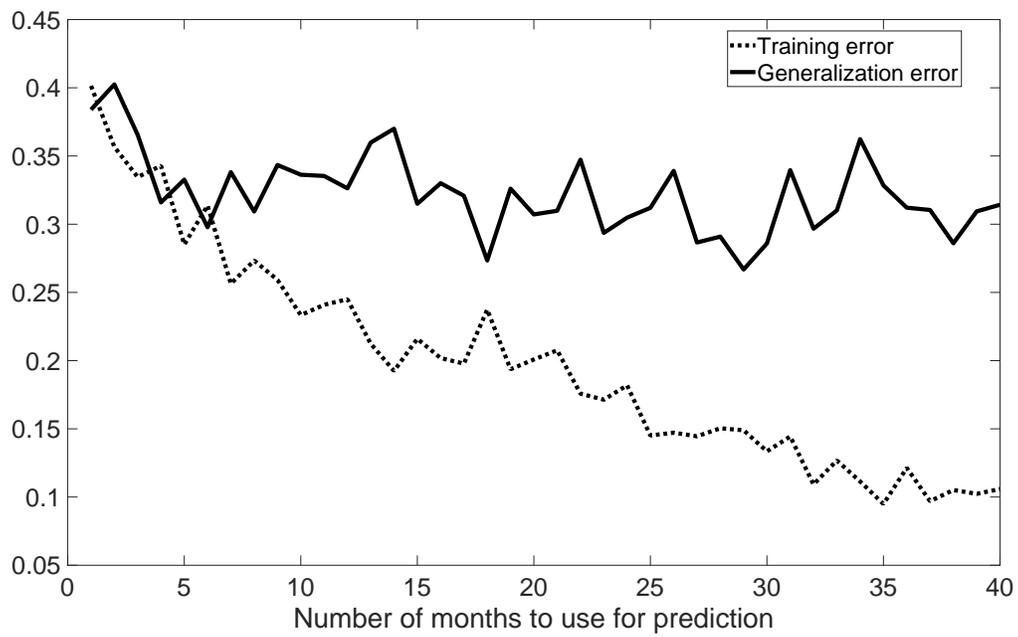}
  \caption{Training error and generalization error of a linear model when changing the number of months to use for prediction. 
  In any setting, a layered neural network achieved better prediction result in terms of the generalization error. }
  \label{fig:appendixC_GT_Linear}
\end{figure}

Here, we experimentally showed that a layered neural network achieved better prediction result than a linear model by using a consumer price index data set in section \ref{sec:exp2}. 
Figures \ref{fig:appendixC_GT_LNN} and \ref{fig:appendixC_GT_Linear}, respectively, show the prediction error of a layered neural network and a linear model. 
These figures show that a layered neural network achieved lower generalization error than a linear model in any number of months to use for prediction. 

\begin{figure}[t]
  \hspace{-15mm}\includegraphics[width=160mm]{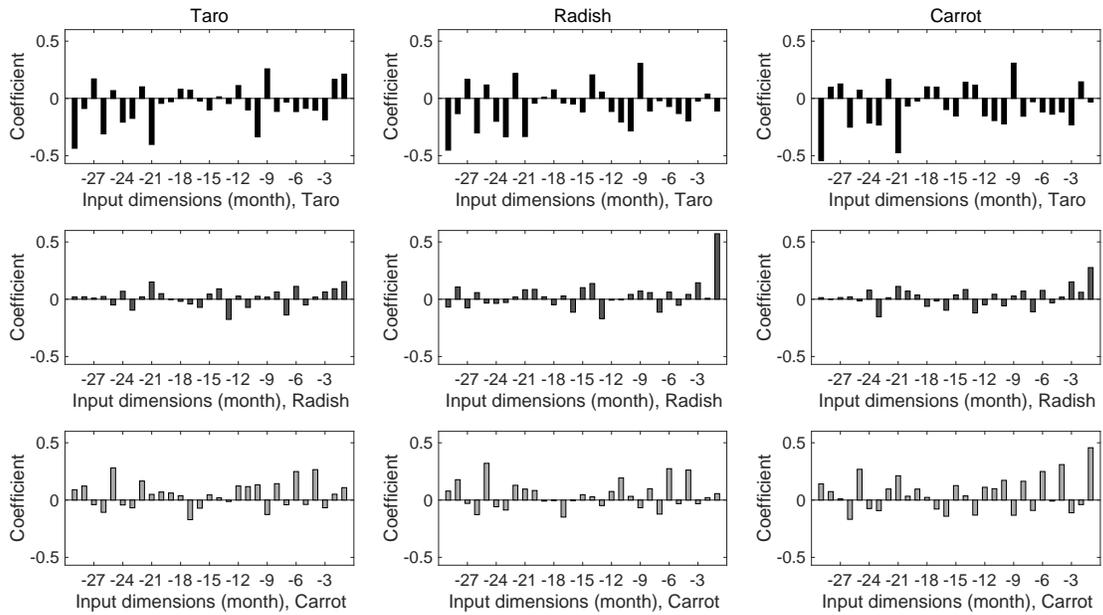}
  \caption{Linear prediction coefficients for a consumer price index data set. 
  This linear model was trained to predict the consumer price indices of taro, radish, and carrot in a month, from the input data of the previous $29$ months of 
  their consumer price indices. }
  \label{fig:appendixC_coef}
\end{figure}

Figure \ref{fig:appendixC_coef} shows the linear prediction coefficients, when setting the number of months to use for prediction at $29$, where the minimum generalization error 
was achieved by a linear model. 



\newpage







\end{document}